\documentclass[twoside,11pt]{article}

\usepackage{jmlr2e}
\usepackage{nicefrac}
\usepackage{amsmath}
\usepackage{textcomp}

\usepackage{subfigure}
\usepackage{enumitem}
\setlist[itemize]{leftmargin=0.8cm}
\setlist[enumerate]{leftmargin=0.8cm}

\usepackage{hyperref}
\hypersetup{hypertexnames=false}
\usepackage{bookmark}
\ShortHeadings{Hidden Talents of the Variational Autoencoder}{Dai, Wang, Aston, Hua and Wipf}
\firstpageno{1}
\input{def.set}

\begin{document}

\title{Hidden Talents of the Variational Autoencoder}

\author{\name Bin Dai \email daib13@mails.tsinghua.edu.cn \\
       \addr Institute for Advanced Study\\
       Tsinghua University\\
       Beijing, China
       \AND
       \name Yu Wang \email yw323@cam.ac.uk \\
       \addr Department of Pure Mathematics and Mathematical Statistics\\
       University of Cambridge\\
       Cambridge, UK
      \AND
      \name John Aston \email j.aston@statslab.cam.ac.uk \\
      \addr Department of Pure Mathematics and Mathematical Statistics\\
      University of Cambridge\\
      Cambridge, UK
      \AND
      \name Gang Hua \email ganghua@microsoft.com\\
      \addr Microsoft Research \\
      Redmond, USA
      \AND
      \name David Wipf \email davidwipf@gmail.com \\
      \addr Microsoft Research \\
      Beijing, China
}

\editor{}

\maketitle

\begin{abstract}%   <- trailing '%' for backward compatibility of .sty file
Variational autoencoders (VAE) represent a popular, flexible form of deep generative model that can be stochastically fit to samples from a given random process using an information-theoretic variational bound on the true underlying distribution.  Once so-obtained, the model can be putatively used to generate new samples from this distribution, or to provide a low-dimensional latent representation of existing samples.  While quite effective in numerous application domains, certain important mechanisms which govern the behavior of the VAE are obfuscated by the intractable integrals and resulting stochastic approximations involved.  Moreover, as a highly non-convex model, it remains unclear exactly how minima of the underlying energy relate to original design purposes.  We attempt to better quantify these issues by analyzing a series of tractable special cases of increasing complexity.  In doing so, we unveil interesting connections with more traditional  dimensionality reduction models, as well as an intrinsic yet underappreciated propensity for robustly dismissing sparse outliers when estimating latent manifolds.  With respect to the latter, we demonstrate that the VAE can be viewed as the natural evolution of recent robust PCA models, capable of learning nonlinear manifolds of unknown dimension obscured by gross corruptions.  A version of this work has appeared in the Journal of Machine Learning Research (JMLR) \cite{dai2018jmlr}; however, we include several small updates here.
%However, this and other previously unexplored talents come at the cost of potential model collapse to a degenerate distribution that may be less suitable as the basis for generating new samples.
\end{abstract}


%==========================================================
% introduction
\section{Introduction}
\label{sec:intro}
\vspace*{-0.2cm}
We begin with a dataset $\bX = \{ \bx^{(i)} \}_{i=1}^n$ composed of $n$ i.i.d.~samples of some random variable $\bx \in \mathbb{R}^d$ of interest, with the goal of estimating a tractable approximation for $p_{\tiny \btheta}(\bx)$, knowledge of which would allow us to generate new samples of $\bx$.  Moreover we assume that each sample is governed by unobserved latent variables $\bz \in \mathbb{R}^{\kappa}$, such that $p_{\tiny \btheta}(\bx) = \int p_{\tiny \btheta} (\bx | \bz) p(\bz) d \bz$, where $\btheta$  are the parameters defining the distribution we would like to estimate.

Given that this integral is intractable in all but the simplest cases, variational autoencoders (VAE) represent a powerful means of optimizing with respect to $\btheta$ a tractable upper bound on $-\log p_{\tiny \btheta} (\bx)$ \cite{Kingma2014,Rezende2014}.  Once these parameters are obtained, we can then generate new samples from $p_{\tiny \btheta}(\bx)$ by first drawing some $\bz^{(i)}$ from $p(\bz)$, and then a new $\bx^{(i)}$ from $p_{\tiny \btheta} (\bx | \bz^{(i)})$.  The VAE upper bound itself is constructed as
\begin{equation}
\hspace*{-0.6cm} \calL(\btheta,\bphi) ~=~  \sum_i \left\{ \mathbb{KL}\left[ q_{\tiny \bphi}\left(\bz|\bx^{(i)} \right) || p_{\tiny \btheta} \left( \bz | \bx^{(i)} \right) \right] -\log p_{\tiny \btheta} (\bx^{(i)}) \right\} ~\geq~ - \sum_i \log p_{\tiny \btheta} (\bx^{(i)}),
\end{equation}
where $q_{\tiny \bphi}\left(\bz|\bx^{(i)} \right)$ defines an arbitrary approximating distribution, parameterized by $\bphi$, and $\mathbb{KL}\left[ \cdot || \cdot \right]$ denotes the KL divergence between two distributions, which is always a non-negative quantity.  For optimization purposes, it is often convenient to re-express this bound as
\begin{equation} \label{eq:vae_cost}
\hspace*{-0.8cm} \calL(\btheta,\bphi) ~\equiv~  \sum_i \left( \mathbb{KL}\left[ q_{\tiny \bphi}\left(\bz|\bx^{(i)} \right) || p(\bz) \right] - \mathbb{E}_{q_{\tiny \bphi}\left(\bz|\bx^{(i)} \right)} \left[\log p_{\tiny \btheta} \left(\bx^{(i)} | \bz  \right)  \right] \right).
\end{equation}
In these expressions, $q_{\tiny \bphi}\left(\bz|\bx \right)$ can be viewed as an \emph{encoder} model that defines a conditional distribution over the latent `code' $\bz$, while $p_{\tiny \btheta} \left(\bx | \bz  \right)$ can be interpreted as a \emph{decoder} model since, given a code $\bz$ it quantifies the distribution over $\bx$.

By far the most common distributional assumptions are that $p(\bz) = \calN(\bz; {\bf 0}, \bI)$ and the encoder model satisfies $q_{\tiny \bphi}\left(\bz|\bx \right) = \calN(\bz; \bmu_z, \bSigma_z)$, where the mean $\bmu_z$ and covariance $\bSigma_z$ are some function of model parameters $\bphi$ and the random variable $\bx$.  Likewise, for the decoder model we assume $p_{\tiny \btheta}\left(\bx|\bz \right) = \calN(\bx; \bmu_x, \bSigma_x)$ for continuous data, with means and covariances defined analogously.\footnote{For discrete data, a Bernoulli distribution is sometimes adopted instead.}

For arbitrarily parameterized moments $\bmu_z$, $\bSigma_z$, $\bmu_x$, and $\bSigma_x$, the KL divergence in (\ref{eq:vae_cost}) computes to
\begin{equation} \label{eq:KL_term}
2 \mathbb{KL}\left[ q_{\tiny \bphi}\left(\bz|\bx \right) || p(\bz) \right] ~\equiv~ \mbox{tr}\left[\bSigma_z \right] + \| \bmu_z \|_2^2 - \log \left| \bSigma_z \right|,
\end{equation}
excluding irrelevant constants.  However, the remaining integral from the expectation term admits no closed-form solution, making direct optimization over $\btheta$ and $\bphi$ intractable.  Likewise, any detailed analysis of the underlying objective function becomes problematic as well.

At least for practical purposes, one way around this is to replace the troublesome expectation with a Monte Carlo stochastic approximation \cite{Kingma2014,Rezende2014}.  More specifically we utilize
\begin{equation} \label{eq:monte_carlo_approx}
\mathbb{E}_{q_{\tiny \bphi}\left(\bz|\bx^{(i)} \right)} \left[\log p_{\tiny \btheta} \left(\bx^{(i)} | \bz  \right)  \right] ~\approx~ \tfrac{1}{\tau} \sum_{t=1}^{\tau} \log p_{\tiny \btheta} \left(\bx^{(i)} | \bz^{(i,t)}  \right),
\end{equation}
where $\bz^{(i,t)}$ are samples drawn from $q_{\tiny \bphi}\left(\bz|\bx^{(i)} \right)$.  Using a simple reparameterization trick, these samples can be constructed such that gradients with respect to $\bmu_z$ and $\bSigma_z$ can be propagated through the righthand side of (\ref{eq:monte_carlo_approx}).  Therefore, assuming all the required moments $\bmu_z$, $\bSigma_z$, $\bmu_x$, and $\bSigma_x$ are differentiable with respect to $\bphi$ and $\btheta$, the entire model can be updated using SGD \cite{Bottou2010}.

While quite effective in numerous application domains that can apply generative models, e.g., semi-supervised learning \cite{Kingma2014b,Maaloe2016,Mansimov2016}, certain important mechanisms which dictate the behavior of the VAE are obfuscated by the required stochastic approximation and the opaque underlying objective with high-dimensional integrals.  Moreover, it remains unclear to what extent minima remain anchored at  desirable locations in the non-convex energy landscape.

We take a step towards better quantifying such issues by probing the basic VAE model under a few simplifying assumptions of increasing complexity whereby closed-form integrations are (partially) possible.  This process unveils a number of interesting connections with more transparent, established generative models, each of which shed light on how the VAE may perform under more challenging conditions.  This mirrors the rich tradition of analyzing deep networks under various simplifications such as linear layers or i.i.d.~random activation patterns \cite{Choromanska2015a,Choromanska2015b,Goodfellow2016,Kawaguchi2016,Saxe2014}, and results in the following key contributions:

\begin{enumerate} % [fullwidth,itemindent=0.5cm]

\item We demonstrate that the canonical form of the VAE, including the Gaussian distributional assumptions described above, harbors an innate agency for robust outlier removal in the context of learning inlier points constrained to a manifold of unknown dimension.  In fact, when the decoder mean $\bmu_x$ is restricted to an affine function of $\bz$, we prove that the VAE model collapses to a form of robust PCA (RPCA) \cite{Candes11,Chandrasekaran11}, a recently celebrated technique for separating data into low-rank (low-dimensional) inlier and sparse outlier components.\footnote{RPCA represents a rather dramatic departure from vanilla PCA and is characterized by a challenging, combinatorial optimization problem.  A formal definition will be provided in Section \ref{sec:rpca}.}

    % However, the VAE maintains noteworthy advantages, such as a natural extensibility to handle outlier-robust low-dimensional manifold learning.

\item We elucidate two central, albeit underappreciated roles of the VAE encoder covariance $\bSigma_z$.  First, through subtle multi-tasking efforts in both terms of (\ref{eq:vae_cost}), it facilitates learning the correct inlier manifold dimension. Secondly, $\bSigma_z$ can help to smooth out undesirable minima in the energy landscape of what would otherwise resemble a more traditional deterministic autoencoder (AE) \cite{Bengio2009}.  This is true even in certain situations where it provably does not actually alter the globally optimal solution itself.  Note that prior to this work the AE could ostensibly be viewed as the most natural candidate for instantiating extensions of RPCA to handle outlier-robust nonlinear manifold learning.  However, our results suggest that the VAE maintains pivotal advantages in mitigating the effects of bad local solutions and over-parameterized latent representations, even in completely deterministic settings that require no generative model per se.

    % The suggests a pivotal, previously unknown advantage of the VAE over the AE, that latter of which, at least prior to our work, could ostensibly be viewed as a natural candidate for instantiating non-affine extensions of robust PCA. % cousin ... even if at times it may not alter the globally optimal solution, or contribute to sample diversity when operating as a generative model.
\end{enumerate}
\noindent As we will soon see, these points can have profound practical repercussions in terms of how VAE models are interpreted and deployed.  For example, one immediate consequence is that even if the decoder capacity is not sufficient to capture the generative distribution \emph{within} some fixed, unknown manifold, the VAE can nonetheless still often find the correct manifold itself, which is sufficient for deterministic recovery of uncorrupted inlier points.  This is exactly analogous to RPCA recovery results, whereby it is possible to correctly estimate an unknown low-dimensional linear subspace heavily corrupted with outliers even if in doing so we do not obtain an actual generative model for the inliers within this subspace.  We emphasize that this is \emph{not} a job description for which the VAE was originally motivated, but a useful hidden talent nonetheless.

%\footnote{By finding the correct manifold, we simply mean that a low-dimensional latent representation is discovered such that the training data $\bX$, excluding any outliers, can be perfectly represented, excluding outliers.}

The remainder of this paper is organized as follows.  In Section \ref{sec:ppca} we consider two affine decoder models and connections with past probabilistic PCA-like approaches. Note that the seminal work from \cite{Rezende2014} mentions in passing that a special case of their VAE decoder model reduces to factor analysis \cite{Bartholomew1999}, a cousin of probabilistic PCA; however, no rigorous, complementary analysis is provided, such as how latent-space sparsity can emerge as we will introduce shortly.  Next we examine various partially affine decoder models in Section \ref{sec:rpca}, whereby only the mean $\bmu_x$ is affine while $\bSigma_x$ has potentially unlimited complexity; all encoder quantities are likewise unconstrained.  We precisely characterize how minimizers of the VAE cost, although not available in closed form, nonetheless are capable of optimally decomposing data into low-rank and sparse factors akin to RPCA while avoiding bad local optima.  This section also discusses extensions as well as interesting behavioral properties of the VAE.

Section~\ref{sec:degeneracy} then considers degeneracies in the full VAE model that can arise even with a trivially simple encoder and corresponding latent representation. Section \ref{sec:experiments} concludes with experiments that directly corroborate a number of interesting, practically-relevant hypotheses generated by our theoretical analyses, suggesting novel usages (unrelated to generating samples) as a tool for deterministic manifold learning in the presence of outliers.  We provide final conclusions in Section~\ref{sec:discussion}.  Note that our prior conference paper has presented the basic demonstration that VAE models can be applied to tackling generalized robust PCA problems \cite{Wang2017}.  However this work primarily considers empirical demonstrations and high-level motivations, with minimal analytical support.

\textbf{Notation:}  We use a superscript $^{(i)}$ to denote quantities associated with the $i$-th sample, which at times may correspond with the columns of a matrix, such as the data $\bX$ or related.  For a general matrix $\bM$, we refer to the $i$-th row as $\bm_{i \cdot}$ and the $j$-th column as $\bm_{ \cdot j}$.  Although technically speaking posterior moments are functions of the parameters $\{\btheta, \bphi\}$, the random variables $\bx$, and the latent $\bz$, i.e., $\bmu_x \equiv \bmu_x\left(\bz; \btheta  \right)$, $\bSigma_x \equiv \bSigma_x\left(\bz; \btheta  \right)$, $\bmu_z \equiv \bmu_z\left(\bx; \bphi  \right)$, and $\bSigma_z \equiv \bSigma_z\left(\bx; \bphi  \right)$, except in cases where some ambiguity exists regarding the arguments, these dependencies are omitted to avoid undue clutter; likewise for $\bmu^{(i)}_z \triangleq \bmu_z(\bx^{(i)} ; \bphi)$ and $\bSigma^{(i)}_z \triangleq \bSigma_z(\bx^{(i)} ; \bphi)$.   Also, with some abuse of notation, we will use $\calL$ to denote a number of different VAE-related objective functions and bounds, with varying arguments and context serving as differentiating factors.  Finally, the $\mbox{diag}[\cdot]$ operator converts vectors to a diagonal matrix, and vice versa as in the Matlab computing environment.

%===================================================
\section{Affine Decoder and Probabilistic PCA}
\label{sec:ppca}

If we assume that $\bSigma_x$ is fixed at some $\lambda \bI$, and force $\bSigma_z = {\bf 0}$ (while removing the now undefined $\log|\bSigma_z|$ term), then it is readily apparent that the resultant VAE model reduces to a traditional AE with squared-error loss function \cite{Bengio2009}, a common practical assumption. To see this, note that if $\bSigma_z = {\bf 0}$, then $q_{\tiny \bphi}\left(\bz|\bx^{(i)} \right)$ collapses to $\delta(\bmu_z)$, i.e., a delta function at the posterior mean, and $\mathbb{E}_{q_{\tiny \bphi}\left(\bz|\bx^{(i)} \right)} \left[\log p_{\tiny \btheta} \left(\bx^{(i)} | \bz  \right)  \right] = \log p_{\tiny \btheta} \left(\bx^{(i)} | \bmu_z^{(i)}  \right)$, which is just a standard AE with quadratic loss and representation $\bmu_x\left( \bmu_z\left[\bx \right] \right)$.  Moreover, the only remaining (non-constant) regularization from the KL term is $\sum_i \| \bmu_z^{(i)}\|_2^2$.  However, given scaling ambiguities that may arise in the decoder when $\bSigma_z = {\bf 0}$, $\bmu_z^{(i)}$ can often be made arbitrarily small, and therefore the effect of this quadratic penalty is infinitesimal.  With affine encoder and decoder models, the resulting deterministic network will simply learn principal components like vanilla PCA, a well-known special case of the AE \cite{Bourlard1988}.

Therefore to understand the VAE, it is crucial to explore the role of non-trivial selections for the encoder and decoder covariances, that serve as both enlightening and differentiating factors.  As a step in this direction, we will explore several VAE reductions that lead to more manageable (yet still representative) objective functions and strong connections to existing probabilistic models.  In this section we begin with the following simplification:

\begin{lemma} \label{lem:ppca}
Suppose that the decoder moments satisfy $\bmu_{x} = \bW \bz + \bb$ and $\bSigma_{x} = \lambda \bI$ for some parameters $\btheta = \{\bW, \bb, \lambda \}$ of appropriate dimensions.  Furthermore, we assume for the encoder we have $\bmu_z = f(\bx; \bphi)$, $\bSigma_z = \bS_z \bS_z^{\top}$, and $\bS_z = g(\bx; \bphi)$, where $f$ and $g$ are any parameterized functional forms that include arbitrary affine transformations for some arrangement of parameters.  Under these assumptions, the objective from (\ref{eq:vae_cost}) admits optimal, closed-form solutions for $\bmu_z$ and $\bSigma_z$ in terms of $\bW$, $\bb$, and $\lambda$ such that the resulting VAE cost collapses to
\begin{equation} \label{eq:ppca_cost}
\calL(\bW, \bb, \lambda) ~=~  \sum_i \bOmega^{(i)} (\bW,\bb,\lambda \bI) + n \log \left| \lambda \bI + \bW \bW^{\top}   \right|,
\end{equation}
where
\begin{equation}
\bOmega^{(i)}(\bW,\bb,\bPsi)~\triangleq~ \left( \bx^{(i)} - \bb \right)^{\top} \left(\bPsi + \bW \bW^{\top}   \right)^{-1} \left( \bx^{(i)} - \bb \right).
\end{equation}
Additionally, if we enforce that off-diagonal elements of $\bSigma_z$ must be equal to zero (i.e., $\left[\bSigma_z\right]_{ij} = 0$ for $i \neq j$), then (\ref{eq:ppca_cost}) further decouples/separates to
\begin{equation} \label{eq:ppca_cost_separable}
\hspace*{-1.3cm} \calL_{\tiny sep}(\bW, \bb, \lambda) ~=~  \sum_i \bOmega^{(i)}(\bW,\bb,\lambda \bI) +  n \left[ \sum_j \log \left(\lambda + \|\bw_{\cdot j} \|_2^2  \right) + (d-\kappa)\log \lambda \right] .
\end{equation}
\end{lemma}

All proofs are deferred to the appendices.  The objective (\ref{eq:ppca_cost}) is the same as that used by certain probabilistic PCA models \cite{Tipping1999}, although the latter is originally derived in a completely different manner.  Moreover, it can be shown that any minimum of this objective represents a globally optimal solution (i.e, no minima with suboptimal objective function value exist).  And with $\bb$ and $\lambda$ fixed, the optimal $\bW$ will be such that $\mbox{span}[\bW]$ equals the span of the singular vectors of $\bX - \bb {\bf 1}^{\top}$ associated with singular values greater than $\sqrt{\lambda}$.  So the global optimum produces a principal subspace formed by soft-thresholding the singular values of $\bX - \bb {\bf 1}^{\top}$, with the rank one offset typically used to simply normalize samples to have zero mean, which is optimal for both (\ref{eq:ppca_cost}) and (\ref{eq:ppca_cost_separable}) alike.\footnote{While the details are omitted here, an optimal solution for $\lambda$ exists in closed form as well.}

In contrast, the alternative cost (\ref{eq:ppca_cost_separable}), which arises from the oft-used practical assumption that $\bSigma_z$ is diagonal, represents a rigorous upper bound to (\ref{eq:ppca_cost}), since
\begin{equation}
 \sum_j \log \left(\lambda + \|\bw_{\cdot j} \|_2^2  \right) + (d-\kappa)\log \lambda ~\geq~ \log \left| \lambda \bI + \bW \bW^{\top}   \right|
\end{equation}
by virtue of Hadamard's inequality (see proof of Theorem \ref{thm:ppca_minima} below), with equality iff $\bW^{\top} \bW $ is diagonal.  Interestingly, all minima of the modified cost nonetheless retain global optimality of the original; however, it can be shown that there will be a combinatorial increase in the actual number of distinct (disconnected) minima:\footnote{By disconnected we mean that, to traverse from one minimum to another, we must ascend the objective function at some point along the way.}
\begin{theorem} \label{thm:ppca_minima}
Let $\bR \in \mathbb{R}^{\kappa \times \kappa}$ denote an arbitrary rotation matrix and $\bP  \in \mathbb{R}^{\kappa \times \kappa}$ an arbitrary permutation matrix.  Furthermore let $\bW^*$ be a minimum of (\ref{eq:ppca_cost}) and $\bW^{**}$ any minimum of (\ref{eq:ppca_cost_separable}) with $\bb$ and $\lambda$ fixed.  Then the following three properties hold:
\begin{enumerate}
\item $\calL(\bW^*, \bb, \lambda)  =  \calL(\bW^* \bR, \bb, \lambda) = \calL_{\tiny sep}(\bW^{**}, \bb, \lambda)$ \begin{equation}\label{eq:ppca_equal_cost}
 = ~~ \calL(\bW^{**} \bP, \bb, \lambda)  =  \calL_{\tiny sep}(\bW^{**} \bP, \bb, \lambda).
\end{equation}
\item For any $\bW^{**} \left(\bW^{**} \right)^{\top}$ with distinct nonzero eigenvalues, there will exist at least $\frac{\kappa!}{\left( \kappa - r \right)!}$ distinct (disconnected) minima of (\ref{eq:ppca_cost_separable}) located at some $\bU \bLambda \bP$, where $\bU \bLambda^2 \bU^{\top}$ represents the SVD of~~$\bW^{**} \left(\bW^{**} \right)^{\top}$ and $r = \mbox{rank}\left[ \bW^{**} \right]$.
 \item $\bW^{**}$ will have at most $r$ nonzero columns, while $\bW^*$ can have any number in $\{r,\ldots,\kappa\}$.
\end{enumerate}
\end{theorem}
\vspace*{0.2cm}
And in terms of local minima, we also have the following:
\begin{corollary} \label{cor:affine_vae_model}
%\vspace{-0.2cm}
For any fixed $\bb$ and $\lambda$, all local minima of (\ref{eq:ppca_cost_separable}) are also global minima.
\end{corollary}
\vspace*{0.1cm}

Although these results\footnote{Corollary \ref{cor:affine_vae_model} is the primary technical result presented herein that is not also contained in \cite{dai2018jmlr}.  Instead, we originally included it within an alternative submission to ICML 2019 (that was not ultimately accepted).  A related result has been independently proposed in a recent ICLR 2019 workshop paper \cite[Theorem 1]{lucas2019understanding}, although the context and proof are quite different.  Regardless, we believe reference \cite{lucas2019understanding} to be complementary to our own.} apply to relatively simplistic affine decoders (the encoder need not be so constrained however), it nonetheless highlights a couple interesting principles.  First, the diagonalization of $\bSigma_z$ collapses the space of globally minimizing solutions to a subset of the original.  While the consequences of this may be minor in the fully affine decoder model where all the minima (global or local) are still equally good, we surmise that with more sophisticated parameterizations this partitioning of the energy landscape into distinct basins-of-attraction could potentially introduce suboptimal local extrema.  And from a broader perspective, Theorem \ref{thm:ppca_minima} and Corollary \ref{cor:affine_vae_model} provide tangible validation of prior conjectures that variational Bayesian factorizations of this sort can fragment the space of local minima \cite{Hoffman2014}.

But there is a second, potentially-advantageous counter-affect elucidated by Theorem \ref{thm:ppca_minima} as well.  Specifically, even if $\bW$ is overparameterized, meaning that $\kappa$ is unnecessarily large, there exists an inherent mechanism to prune superfluous columns to exactly zero, i.e., column-wise sparsity.  And once columns of $\bW$ become sparse, the corresponding elements of $\bmu_z$ can no longer influence the data fit.  Consequently, the $\|\bmu_z\|^2_2$ factor from (\ref{eq:KL_term}) serves as the only relevant influence, pushing these values to be exactly zero even though $\ell_2$ norms in most regularization contexts tend to favor diverse, \emph{non-sparse} representations \cite{Rao03}.

So ultimately, sparsity of $\bmu_z$ in this context is an artifact of the diagonal $\bSigma_z$ assumption and the interaction of multiple VAE terms, a subtle influence we empirically demonstrate translates to more complex regimes in Section \ref{sec:experiments}.  In any event, we have shown that both variants of the affine decoder model lead to reasonable probabilistic PCA-like objectives regardless of how overparameterized $\bmu_z$ and $\bSigma_z$ happen to be.

%=====================================================
\section{Partially Affine Decoder and Robust PCA}
\label{sec:rpca}

Thus far we have considered tight limitations on the complexity allowable in the functional forms of both $\bmu_x$ and $\bSigma_x$, while $\bmu_z$ and $\bSigma_z$ were free-range variables granted arbitrary flexibility. We now turn our gaze to the case where $\bSigma_x$ can also be any parameterized, diagonal matrix\footnote{A full covariance over $\bx$ is infeasible given the high dimension, and can lead to undesirable degeneracies anyway.  Therefore a diagonal covariance is typically, if not always, used in practice.} while $\bmu_x$ remains restricted.  Although this administers considerable capacity to the model at the potential risk of overfitting, we will soon see that the VAE is nonetheless able to self-regularize in a very precise sense: Global minimizers of the VAE objective will ultimately correspond with optimal solutions to
\begin{equation} \label{eq:canonical_rpca}
\min_{\bL,\bS} ~~~ n \cdot \mbox{rank}\left[ \bL \right] + \| \bS \|_0, ~~~~ \mbox{s.t. } \bX = \bL + \bS,
\end{equation}
where $\| \cdot \|_0$ denotes the $\ell_0$ norm, or a count of the number of nonzero elements in a vector or matrix.  This problem represents the canonical form of robust principal component analysis (RPCA) \cite{Candes11,Chandrasekaran11}, decomposing a data matrix $\bX$ into low-rank principal factors $\bL = \bU \bV$, with $\bU$ and $\bV$ low-rank matrices of appropriate dimension, and a sparse outlier component $\bS$.  However, we must emphasize that (\ref{eq:canonical_rpca}), unlike traditional PCA, represents an NP-hard, discontinuous optimization problem with a combinatorial number of potentially bad local minima.  Still, it is seemingly quite remarkable that the probabilistic VAE model shares any kinship with (\ref{eq:canonical_rpca}), even more so given that some of the distracting \emph{local} minimizers can be smoothed away, a key VAE advantage as we will later argue.

%In fact, if we were to replace the $\ell_0$ penalty with an $\ell_2$ norm, we would exactly recover one formulation of vanilla PCA, where by replacing the weighting factor $n$ with different arbitrary scalars, the optimal solution will simply reflect principal subspaces of varying dimensions.  That the complex, probabilistic VAE model shares any kinship with (\ref{eq:canonical_rpca}) is seemingly quite remarkable.

Before elucidating this relationship, we require one additional technical caveat.  Specifically, since $\log 0$ and $\tfrac{1}{0}$ are both undefined, and yet we will soon require an alliance with degenerate (or nearly so) covariance matrices that mimic the behavior of sparse and low-rank factors through log-det and inverse terms, we must place the mildest of restrictions on the minimal allowable singular values of $\bSigma_x$ and $\bSigma_z$.  For this purpose we define $\calS^m_\alpha$ as the set of $m \times m$ covariance matrices with singular values all greater than or equal to $\alpha$, and likewise $\bar{\calS}^m_\alpha$ as the subset of $\calS^m_\alpha$ containing only diagonal matrices.  We also define  $\mbox{supp}_{\alpha}( \bx ) = \{i : |x_i| > \alpha\}$, noting that per this definition, $\mbox{supp}_{0}( \bx ) = \mbox{supp}( \bx )$, meaning we recover the standard definition of support: the set of indices associated with nonzero elements.

\subsection{Main Result and Interpretation}

Given the affine assumption from above, and the mild restriction $\bSigma_x \in \bar{\calS}^d_{\alpha}$ and $\bSigma_z \in \calS^{\kappa}_{\alpha}$ for some small $\alpha > 0$, the resulting constrained VAE minimization problem can be expressed as
\begin{equation}  \label{eq:vae_rpca_global_min}
\min_{\btheta,\bphi} ~~ \calL\left(\bW,\bb  = {\bf 0}, \bSigma_x \in \bar{\calS}^d_{\alpha}, \bmu_z, \bSigma_z \in \calS^{\kappa}_{\alpha} \right),
\end{equation}
where now $\btheta$ includes $\bW$ as well as all the parameters embedded in $\bSigma_x$, while $\bmu_z$ and $\bSigma_z$ are parameterized as in Lemma \ref{lem:ppca}.  We have also set $\bb = {\bf 0}$ merely for ease of presentation as its role is minor. We then have the following:

\begin{theorem} \label{thm:vae_and_rpca}
Suppose that $\bX = \{ \bx^{(i)} \}_{i=1}^n$ admits a feasible decomposition $\bX = \bU \bV + \bS$ that uniquely\footnote{Obviously only $\bL$ and $\bS$ will be unique; the actual decomposition of $\bL$ into $\bU$ and $\bV$ is indeterminate up to an inconsequential invertible transform.} optimizes (\ref{eq:canonical_rpca}).  Then for some $\bar{\alpha}$ sufficiently small, and all $\alpha \in (0,\bar{\alpha} ]$, any global minimum $\{ \hat{\bW}, \hat{\bSigma}_x, \hat{\bmu}_z, \hat{\bSigma}_z \}$ of (\ref{eq:vae_rpca_global_min})  will be such that\footnote{Although somewhat cumbersome in print, the expression $\hat{\bSigma}_x\left(\hat{\bmu}_z\left[ \bx^{(i)}\right]\right)$ refers to $\hat{\bSigma}_x$ evaluated at $\hat{\bmu}_z$, where the latter is evaluated at $\bx^{(i)}$, the $i$-th sample.}
\begin{equation} \label{eq:optimal_conditions}
\mbox{span}[\hat{\bW}]  =  \mbox{span}[\bU]  ~~~~~\mbox{and} ~~~~~ \mbox{supp}_{\alpha}\left( \mbox{diag}\left[\hat{\bSigma}_x\left(\hat{\bmu}_z\left[ \bx^{(i)}\right]\right)\right] \right) = \mbox{supp}[\bs^{(i)}]
\end{equation}
for all $i$ provided that the latent representation satisfies $\kappa \geq \mbox{rank}\left[\bU\right]$.
\end{theorem}

Several important remarks are warranted here regarding the consequences and interpretation of this result:
\begin{itemize}
\item The $\hat{\bW}$ satisfying (\ref{eq:optimal_conditions}) forms a linear basis for each inlier component $\bl^{(i)}$, and likewise, a sample-dependent basis denoted $\bE^{(i)}$ can be trivially constructed for each outlier component $\bs^{(i)}$ using $\hat{\bSigma}_x$, and $\hat{\bmu}_z$.  Specifically, each unique column of $\bE^{(i)}$ is a vector of zeros with a one in the $j$-th position, with $j \in \mbox{supp}_{\alpha}\left( \mbox{diag}\left[\hat{\bSigma}_x\left(\hat{\bmu}_z\left[ \bx^{(i)}\right]\right)\right] \right)$.  It follows that
    \vspace*{-0.3cm}
    \begin{equation}
    \bx^{(i)} = \bl^{(i)} + \bs^{(i)} = \left[ \hat{\bW} ~~ \bE^{(i)}\right] \left[ \hat{\bW} ~~ \bE^{(i)}\right]^{\dag} \bx^{(i)}, ~~ \forall i = 1,\ldots,n.
    \end{equation}
Therefore if we can globally optimize the VAE objective, we can recover the correct latent representation, or equivalently, the optimal solution to (\ref{eq:canonical_rpca}).

%Given $\hat{\bW}$, $\hat{\bSigma}_x$, and $\hat{\bmu}_z$ satisfying (\ref{eq:optimal_conditions}) as described, we can directly recover the generating low-rank $\bL = \bU \bV$ and sparse $\bS$ by solving a simple linear system for each data point $i$.  See Appendix \ref{sec:optimal_solution_computations} for the details.  Therefore if we can globally optimize the VAE objective, we can recover the correct latent representation, or equivalently, the optimal solution to (\ref{eq:canonical_rpca}).
%
%$\bE^{(i)}$ is a matrix of zeros with a single `1' in each column.  Additionally, $\bE^{(i)}$ has a separate column for every $j \in \mbox{supp}_{\alpha}\left( \mbox{diag}\left[\hat{\bSigma}_x\left(\hat{\bmu}_z\left[ \bx^{(i)}\right]\right)\right] \right)$, in each case constructed with a `1' located in the $j$-th position and zeros elsewhere.
%
% Basically, (\ref{eq:vae_decomp}) implies that we are representing each observed $\bx^{(i)}$ as an inlier component $\hat{\bW} \bomega_1$ and an outlier component $\bE^{(i)} \bomega_2$.

\item The requirements $\bSigma_x \in \bar{\calS}^d_{\alpha}$ and $\bSigma_z \in \calS^{\kappa}_{\alpha}$ do not portend the need for specialized tuning or brittleness of the result; these are merely technical conditions for dealing with degenerate covariances that occur near optimal solutions.  While it might seem natural that $\bSigma_x$ has diagonal elements pushed to zero in regions where near perfect data fit is possible, less intuitively, global optima of (\ref{eq:vae_rpca_global_min}) can be achieved with an arbitrarily small $\bSigma_z$, e.g., $\bSigma_z = \alpha \bI$,  at least along latent dimensions needed to represent $\bL$ (see proof construction).  And interestingly, this implies that in areas surrounding a global optimum, the VAE objective can \emph{resemble that of a regular AE}.  As we will discuss more below, desirable smoothing effects of integration over $\bSigma_z$ occur \emph{elsewhere} in the energy landscape while preserving extrema anchored at the correct latent representation.

\item Even if $\kappa$ is large, meaning $\bW$ is possibly overcomplete, the VAE will not overfit in the sense that there exists an inherent regulatory effect pushing $\mbox{span}[\bW]$ towards $\mbox{span}[\bU]$.

\item If the globally optimal solution to (\ref{eq:canonical_rpca}) is not unique (this is different from uniqueness regarding the VAE objective), then a low-rank-plus-sparse model may not be the most reasonable, parsimonious representation of the data to begin with, and exact recovery of $\bL$ and $\bS$ will not be possible by \emph{any} algorithm without further assumptions.  More concretely, an arbitrary data point $\bx^{(i)} \in \mathbb{R}^d$ requires $d$ degrees of freedom  to represent; however, if the data succinctly adheres to the RPCA model, then for properly chosen $\bU$, $\bV$, and $\bS$, we can have $\bx^{(i)} = \bU \bv^{(i)} + \bs^{(i)}$, where $\|\bv^{(i)}\|_0 + \|\bs^{(i)}\|_0 < d$. Arbitrary data in general position will never admit such a unique decomposition, and we should only expect such structure in data well-represented by our VAE model, or the original RPCA predecessor from (\ref{eq:canonical_rpca}).

\item A number of celebrated results have stipulated conditions \cite{Candes11,Chandrasekaran11} whereby global solutions of the convex relaxation into nuclear and $\ell_1$ norm components given by
\begin{equation} \label{eq:pcp}
\min_{\bL,\bS} ~~~ \sqrt{n} \cdot \mbox{rank}\left\| \bL \right\|_* + \| \bS \|_1, ~~~~ \mbox{s.t. } \bX = \bL + \bS,
\end{equation}
will equal global solutions of (\ref{eq:canonical_rpca}).  While elegant in theory, and practically relevant given that (\ref{eq:canonical_rpca}) is discontinuous, non-convex, and difficult to optimize, the required conditions for this equivalence to hold place strong restrictions on the allowable structure in $\bL$ and support pattern in $\bS$.  In practice these conditions can never be verified and are unlikely to hold, so an alternative modeling approach such as the VAE, which can be viewed as a smoothed version of (\ref{eq:canonical_rpca}) when an affine decoder mean is used (more on this later), remains attractive.  Additionally, there is no clear way to modify (\ref{eq:pcp}) to handle nonlinear manifolds, which is obviously the bread and butter of the VAE.

\end{itemize}

We emphasize that these conclusions are not the product of an overly contrived situation, given that a significant restriction is only placed on $\bmu_x$; \emph{all} other posterior quantities are essentially unconstrained provided a sufficient lower complexity bound is exceeded, implying that the result will hold whenever a sufficiently complex deep network is used.  Moreover, although we will defer to a formal treatment to future work for purposes of brevity here, with some mild additional conditions, Theorem \ref{thm:vae_and_rpca} can naturally be extended to the case where the decoder mean function is generalized to subsume non-linear, union-of-subspace models as commonly assumed in subspace clustering problems \cite{Elhamifar13,Rao10}.  This then deviates substantially from any direct PCA-kinship, and buttresses the argument that the analysis presented here transitions to broader scenarios.  The experiments from Section \ref{sec:experiments} will also provide complementary empirical confirmation.

Moving forward, as a point of further comparison it is also interesting to examine how a traditional AE, which emerges when $\bSigma_z$ is forced to zero, behaves under analogous conditions to Theorem \ref{thm:vae_and_rpca}.
\begin{corollary} \label{cor:AE_behavior}
Under the same conditions as Theorem \ref{thm:vae_and_rpca}, if we remove the $\log \left| \bSigma_z \right|$ term and assume $\bSigma_z = {\bf 0}$ elsewhere, then (\ref{eq:vae_rpca_global_min}) admits a closed-form solution for $\bSigma_x$ in terms of $\bW$ and $\bmu_z$ such that minimizers of the VAE cost are minimizers of
\begin{equation} \label{eq:canonical_rpca_constrained}
\calL\left(\bW, \bmu_z \right) = \sum_i \left\| \bx^{(i)} - \bW \bmu_z\left(\bx^{(i)} \right) \right\|_0 ~~~ \mbox{ in the limit } \alpha \rightarrow 0.
\end{equation}
\end{corollary}
From this result we immediately observe that, provided $\bmu_z$ enjoys a sufficiently rich parameterization, minimization of (\ref{eq:canonical_rpca_constrained}) is just a constrained version of (\ref{eq:canonical_rpca}), exactly equivalent to solving
\begin{equation} \label{eq:canonical_rpca_constrained2}
\min_{\bL,\bS} ~~~  \| \bS \|_0, ~~~~ \mbox{s.t. } \bX = \bL + \bS,~~ \mbox{rank}\left[ \bL \right] \leq \kappa.
\end{equation}
This expression immediately exposes one weakness of the AE; namely, if $\kappa$ is too large, there is no longer any operation in place to prune away unnecessary dimensions, and the trivial solution $\bL = \bX$ will be produced.  In the large-$\kappa$ regime then, global VAE and global AE solutions do in fact deviate, ultimately because of the removal of the $-\log|\bSigma_z|$ term in the latter.  So $\bSigma_z$ plays a critical role in determining the correct, low-dimensional inlier structure, and ultimately it is this covariance that chaperons $\bW$ during the learning process.

\subsection{Additional Local Minima Smoothing Effects}
There is also a more important, yet subtle, advantage of the VAE over both (\ref{eq:canonical_rpca_constrained2}) and the original unconstrained RPCA model from (\ref{eq:canonical_rpca}).  For both RPCA constructions, any feasible support pattern, even the trivial ones associated with non-interesting decompositions satisfying $\|\bv^{(i)}\|_0 + \|\bs^{(i)}\|_0 \geq d$ for some $i$, will necessarily represent a local minimum, since there is an infinite gradient to overcome to move from a zero-valued element of $\bS$ to a nonzero one.

Unlike these deterministic approaches, the behavior of the VAE reflects a form of differential smoothing that rids the model of many of these pitfalls while retaining desirable minima that satisfy (\ref{eq:optimal_conditions}).\footnote{A more rudimentary form of this smoothing has been observed in much simpler empirical Bayesian models derived using Fenchel duality \cite{wipf2012}.}   Based on details of the proof of Theorem \ref{thm:vae_and_rpca}, it can be shown that, excluding small-order terms dependent on other variables and a constant scale factor of $-\log \alpha$, then a representative bound on the VAE objective associated with each sample index $i$ behaves like
\begin{equation} \label{eq:crude_vae_cost}
\mbox{rank}[\bW] +  \mbox{supp}_{\alpha} \left( \mbox{diag} \left[ \bSigma_x \left(\bmu_z\left[\bx^{(i)}\right]\right) \right] \right).
\end{equation}
But crucially, \emph{this behavior lasts only as long as (\ref{eq:crude_vae_cost}) is strictly less than} $d$ and $\bSigma_z$ is forced to be small or degenerate.  In contrast, when the value is at or above $d$, (\ref{eq:crude_vae_cost}) no longer reflects the energy function, which becomes relatively flat because of smoothing via $\bSigma_z$, avoiding the pitfalls described above.  This phenomena then has the potential to smooth out a large constellation of bad locally optimal solutions.

To situate things in the narrative of (\ref{eq:canonical_rpca}), which is useful for illustration purposes, the VAE can be viewed (at least to first order approximation) as minimizing the alternative lower-bounding objective function
\begin{eqnarray}
\sum_i \mbox{rank}\left[\bL \bL^{\top} + \mbox{diag}\left( \bs^{(i)} \right)^2   \right] & \leq & \sum_i \mbox{rank}\left[\bL \bL^{\top}\right] + \sum_i \mbox{rank}\left[\mbox{diag}\left( \bs^{(i)} \right)^2   \right] \nonumber \\
& = &  n \cdot \mbox{rank}\left[ \bL \right] + \left\| \bS \right\|_0,
\end{eqnarray}
or a smooth surrogate thereof, over the constraint set $\bX = \bL + \bS$.  The advantages of this lower bound are substantial:  As long as a unique solution exists to the RPCA problem, the globally optimal solution with  $\|\bv^{(i)}\|_0 + \|\bs^{(i)}\|_0 < d$ for all $i$ will be unchanged; however, \emph{any} feasible solution with $\|\bv^{(i)}\|_0 + \|\bs^{(i)}\|_0 \geq d$ will have a constant cost via the expression on the left of the inequality, truncating the many erratic peaks that will necessarily occur with the energy on the righthand side.

In fact, away from the strongly attractive basins of optimal VAE solutions, the KL term from (\ref{eq:vae_cost}) is likely to  push $\bSigma_z$ more towards
\begin{equation} \label{eq:sigma_z_identity}
\hspace*{-1.8cm} \arg \min_{\bSigma_z \succ {\bf 0} } \mathbb{KL}\left[ q_{\tiny \bphi}\left(\bz|\bx \right) || p(\bz) \right]  ~~\equiv~~ \arg \min_{\bSigma_z \succ {\bf 0} } \mbox{tr}\left[\bSigma_z \right]  - \log \left| \bSigma_z \right| ~~ = ~~\bI.
\end{equation}
Experiments presented in Section~\ref{sec:experiments} confirm that this is indeed the case.  And once $\bSigma_z$ moves away from zero, it  will generally contribute a strong smoothing effect via the expectation in (\ref{eq:vae_cost}).  However, there exists an important previously unobserved caveat here:  If the decoder mean function is excessively complex, it can potentially outwit all regulatory persuasions from $\bSigma_z$, leading to undesirable degenerate solutions with no representational value as described next.

%===================================================
\section{Degeneracies Arising from a Flexible Decoder Mean}
\label{sec:degeneracy}

In this section we consider the case where $\bmu_x$ is finally released from its affine captivity to join with posterior colleagues in the wild. That simultaneously granting $\bmu_x$, $\bSigma_x$, $\bmu_z$, and $\bSigma_z$ unlimited freedom leads to overfitting may not come as a surprise; however, it turns out that even if the latter three are severely constrained, overfitting will not be avoided when $\bmu_x$ is over-parameterized in a certain sense extending beyond a single affine layer.  This is because, at least at a high level, the once-proud regulatory effects of $\bSigma_z$ can be completely squashed in these situations leading to the following:
\begin{theorem} \label{thm:free_mu_x}
Suppose $\kappa = 1$ (i.e., a latent dimension of only one), $\bSigma_z \equiv \sigma_z^2 = \lambda_z$ (a scalar), $\mu_z = \ba^{\top} \bx$ for some fixed vector $\ba$, $\bSigma_x = \lambda_x \bI$, and $\bmu_x$ is an arbitrary piecewise linear function with $n$ segments.  Then the VAE objective is unbounded from below at a trivial solution $\{\hat{\lambda}_z,\hat{\ba},\hat{\lambda}_x, \hat{\bmu}_x \}$ such that the resulting posterior mean $\hat{\bmu}_x(z;\btheta)$ will satisfy $\hat{\bmu}_x(z; \btheta) \in \{ \bx^{(i)}\}_{i=1}^n$ with probability one for any $z$.
\end{theorem}
In this special case, $\bSigma_x$, $\sigma_z^2$, and $\mu_z$ are all simple affine functions and the latent dimension is minimal, and yet an essentially useless, degenerate solution can arbitrarily optimize the VAE objective.  This occurs because the VAE has limited power to corral certain types of heavily over-parameterized decoder mean functions, even when all other degrees of freedom are constrained, and in this regime the VAE essentially has no advantage over a traditional autoencoder (its natural self-regulatory agency may sometimes break down).  In contrast, as we saw in a previous section, there is no problem taming the influences of an unlimited latent representation (meaning $\kappa$ is large, e.g., even $\kappa > n$) and its huge, attendant parameterized mean function, provided the latter is affine, as in $\bmu_x = \bW \bz + \bb$.

Indeed then, the issue is clearly not the \emph{degree} of over-parameterization in $\bmu_x$ per se, but the actual structures in place.  And the key problem is that, at least in some situations, the model can circumvent the entire regulatory mechanism of the KL term, pushing the latent variances towards zero even around \emph{undesirable} solutions.  For example, in the context of Theorem \ref{thm:free_mu_x}, the piecewise linear structure of $\bmu_x$ allows the decoder to act much like a vector quantization process, encouraging $z$ towards a scalar code that selects for piecewise linear segments matched to training samples $\bx^{(i)}$.  And because this will lead to perfect reconstruction error if an optimal segment is found for a particular $z^{(i)}$, $\bSigma_x = \lambda_x \bI \approx {\bf 0}$ serves as a reasonable characterization of posterior uncertainty, pushing $p(\bx^{(i)} | z^{(i)} ) \rightarrow \delta\left(\bx^{(i)}\right)$ provided that $z^{(i)} \approx \mu_z\left(\bx^{(i)} ; \ba \right) = \ba^{\top} \bx^{(i)}$, meaning that $\sigma_z^2 = \lambda_z$ is not too large.

In this situation, loosely speaking the data term from (\ref{eq:vae_cost}) will behave like $n d \log \lambda_x$, bullying the over-matched KL term that will scale only as $-n \log \lambda_z$.  This in turn leads to a useless, degenerate solution as $\lambda_x = \lambda_z \rightarrow 0$, either for the purposes of generating representative samples, or for outlier removal as we have described herein.

One helpful caveat though, is that actually implementing such a complex piecewise linear function $\hat{\bmu}_x(z;\btheta)$ using typical neural network components would require extremely wide and/or deep structure beyond the first decoder mean layer.  And the degrees of freedom in such higher-layer structures would need to scale proportionally with the size of the training data, which is not a practical VAE operational regime to begin with.  In contrast, the first layer of the decoder mean network more or less self-regularizes, at least in the affine and related cases as described above.  And we conjecture that this self-regularization preserves in more complex networks of reasonable practical size as will be empirically demonstrated in Section \ref{sec:experiments}.  So really it is \emph{excessive} complexity in higher decoder mean layers, unrelated to the dimensionality of the latent $\bz$ bottleneck, where overfitting problems are more likely to arise.

Of course an analogous issue exists with generative adversarial networks (GAN) as well, a popular competing deep generative model composed of a generator network analogous to the VAE decoder, and a discriminator network that replaces the VAE encoder in a loose sense \cite{Goodfellow2014}.  If the generator network merely learns a segmentation of $\bz$-space such that all points in the $i$-th partition map to $\bx^{(i)}$, the discriminator will be helpless to avert this degenerate situation even in principle.  But there is an asymmetry when it comes to the GAN discriminator network and the VAE encoder:  Over-parameterization of the former can be problematic (e.g., it can easily out-wit an affine or other proportionally simple generator), but the latter not so, at least in the sense that a highly flexible VAE encoder need not bully a simple decoder into trivial solutions as we have shown in previous sections.

%===================================================
%\vspace*{-0.2cm}
\section{Experiments and Analysis} \label{sec:experiments}
%\vspace*{-0.2cm}

Theoretical examination of simplified cases can be viewed as a powerful vehicle for generating accessible hypotheses that describe likely behavior in more realistic, practical situations.  In this section we empirically evaluate and analyze three concrete hypotheses that directly emanate from our previous technical results and the tight connections between RPCA and VAE models.  In aggregate, these hypotheses have wide-ranging consequences in terms of how VAEs should be applied and interpreted.

Before stating these hypotheses, we summarize what can be viewed as two, theoretically-accessible boundary cases considered thus far.  First, building on Section \ref{sec:ppca}, Section \ref{sec:rpca} demonstrated that the VAE can self-regularize and produce useful, robust models provided that restrictions are placed on only the decoder mean network.  Conversely, Section \ref{sec:degeneracy} demonstrated that, regardless of other model components, if the decoder mean network is unreasonably complex beyond the first layer, then overfitting emerges as a potential concern.  But between these two extremes, there exists a large operational regime whereby practical VAE behavior is both worth exploring and likely still informed by the original analysis of these boundary cases.

Within this context then, we conjecture that the desirable VAE properties exposed in Sections \ref{sec:ppca} and \ref{sec:rpca} are inherited by models involving deeper decoder mean networks, but at least constrained to practically-sized hidden-layer $\bmu_x$ complexity such that the concerns from Section \ref{sec:degeneracy} are not a significant factor (e.g., no networks where the degrees of freedom in higher decoder mean layers scales as $d\times n$, an absurd VAE structure by any measure).  More specifically, in this section will empirically examine the following three hypotheses:

 % in practice:
\begin{enumerate}[label=(\roman*)]
\item When the decoder mean function is allowed to have multiple hidden layers of sensible size/depth, the VAE should behave like a nonlinear extension of RPCA, but with natural regularization effects in place that help to avoid local minima and/or overfitting to outliers.  It is therefore likely to outperform either RPCA algorithms or, more importantly, an AE on diverse manifold recovery/outlier discovery problems unrelated to the probabilistic generative modeling tasks the VAE was originally designed for.

\item If the VAE latent representation $\bz$ is larger than needed (meaning its dimension $\kappa$ is higher than the true data manifold dimension), we have proven that unnecessary columns of $\bW$ in a certain affine decoder mean model $\bmu_{x} = \bW \bz + \bb$ will automatically be pruned as desired.  Analogously, in the extended nonlinear case we would then expect that columns of the weight matrix from the first layer of the decoder mean network should be pushed to zero, again effectively pruning away the impact of any superfluous elements of $\bz$.\footnote{As opposed to column-sparsity in the first layer weights, it is also conceivable that unnecessary latent dimensions could instead be effectively shut off via more complex nonlinear interactions across multiple layers.  However, this alternative scenario would seemingly be more likely with very deep, high capacity networks and significant coordination would be required.  And at least in our experiments (see below), this did not appear to be occurring.}

\item When granted sufficient capacity in both $\bmu_x\left( \bmu_z\left[\bx \right] \right)$ and $\bSigma_x$ to model inliers and outliers respectively, the VAE should have a tendency to push elements of the encoder covariance $\bSigma_z$ to arbitrarily near zero along latent dimensions needed for representing inlier points, \emph{selectively} overriding the KL regularizer that would otherwise push these values towards one.  This counterintuitive behavior directly facilitates the VAE's utility as a nonlinear outlier removal tool (per Hypothesis (i)) by preserving exact adherence to the manifold in the neighborhood of optimal solutions.
%
%Serve as a bellwether
%
%Although originally promoted as a probabilistic generative model, when granted sufficient capacity in both $\bmu_x\left( \bmu_z\left[\bx \right] \right)$ and $\bSigma_x$ to model inliers and outliers respectively, the VAE should have a tendency to push elements of the encoder covariance $\bSigma_z$ to exactly zero (at least along latent dimensions needed for representing inlier points), completely overriding the KL regularizer that would otherwise push these values towards one.  In doing so, the VAE's value as a true (or non-degenerate) generative model is undermined; however, its utility as a nonlinear outlier removal tool is dramatically enhanced per Hypothesis (i) above.
\end{enumerate}

\subsection{Hypothesis (i) Evaluation Using Specially-Designed Ground-Truth Manifolds}\label{exp:evaluation_i}
 If our theory is generally applicable, then a VAE with suitable parameterization should be able to significantly outperform an analogous deterministic AE (i.e., an equivalent VAE but with $\bSigma_z = {\bf 0}$) on the task of recovering data points drawn from a low-dimensional nonlinear manifold, but corrupted with gross outliers.  In other words, even if both models have equivalent capacity to capture the intrinsic underlying manifold in principle, the VAE is more likely to avoid bad minima and correctly estimate it.  We demonstrate this VAE capability here for the first time across an array of manifold dimensions and corruption percentages, recreating a nonlinear version of what are commonly termed \emph{phase transition plots} in the vast RPCA literature \cite{Candes11,Ding11,oh2013partial,wipf2012}. These plots evaluate the reconstruction quality of competing algorithms for every pairing of subspace dimension and outlier ratio, creating a heat map that differentiates success and failure regions.

%Our goal here is to recreate such phase plots using nonlinear manifolds that better showcase the increased generality of the VAE.

Of course explicit knowledge of ground-truth low-dimensional manifolds is required to accomplish this.  With linear subspaces it is trivial to generate appropriate synthetic data by simply creating two low-rank random matrices $\bU \in \mathbb{R}^{d \times \kappa}$ and $\bV\in \mathbb{R}^{\kappa \times n}$, a sparse outlier matrix $\bS$, and then computing $\bX = \bL + \bS$ with $\bL = \bU \bV$.  Algorithms are presented with only $\bX$ and attempt to reconstruct $\bL$. Here we generalize this process to the nonlinear regime using deep networks and the following non-trivial steps.  In this revised context, the generated $\bL$ will now represent a data matrix with columns confined to a ground-truth nonlinear manifold.

\vspace*{0.5cm}
\textbf{Data Generation:} First we draw $n$ low-dimensional samples $\bz^{(i)} \in \mathbb{R}^{\kappa}$ from $\calN(\bz; {\bf 0},\bI)$ and pass them through a $3$-layer network with ReLU activations \cite{temp2}.  We express this structure as $\bz(\kappa)$-$\bD_1(r_1)$-$\bD_{2}(r_2)$-$\bl(d)$, where $\bD_1$ and $\bD_2$ are hidden layers, $\bl$ here serves as the output layer, and the values inside parentheses denote the respective dimensionalities (these experiment-dependent values will be discussed later).  Network weights are set using the initialization procedure from \cite{he2015delving}.  The $d$-dimensional output produced by $\bz^{(i)}$ is denoted as $\bl^{(i)}$, the collection of which form a matrix $\bL$, with columns effectively lying on a $\kappa$-dimensional nonlinear manifold.  This network can be viewed as a \emph{ground-truth decoder}, projecting $\bz^{(i)}$ to clean samples $\bl^{(i)}$.

But we must also verify that there exists a known \emph{ground-truth encoder} that can correctly invert the decoder, otherwise we cannot be sure that any given VAE structure provably maintains an optimal encoder within its capacity (this is very unlike the linear RPCA case where an analogous condition is trivially satisfied).  To check this, we learn the requisite inverse mapping by training something like an inverted autoencoder.  Basically, the decoder described above now acts as an encoder, to which we append a new $3$-layer ReLU network structured as $\bl(d)$-$\bE_1(r_2)$-$\bE_{2}(r_1)$-$\hat{\bz}(\kappa)$, where now $\bE_1$ and $\bE_2$ denote candidate hidden layers for a potentially optimal encoder.  The entire intverted structure then becomes $\bz(\kappa)$-$\bD_1(r_1)$-$\bD_{2}(r_2)$-$\bl(d)$-$\bE_1(r_2)$-$\bE_{2}(r_1)$-$\hat{\bz}(\kappa)$.  If any $\bz^{(i)}$ passes through this network with zero reconstruction error, it implies that the corresponding $\bl^{(i)}$ can pass through the flipped network with zero reconstruction error, and we have verified our complete ground truth network.

We could train the entire system end-to-end to accomplish this, which should be easy since $\kappa \ll d$; however, we found that although $\bz^{(i)} = \hat{\bz}^{(i)}$ is obviously not difficult to achieve, the corresponding learned samples $\bl^{(i)}$ are pushed to very near a low-rank matrix when assembled into $\bL$.  This would imply that non-linear manifold learning is not actually even required and RPCA would likely be sufficient.

To circumvent this issue, we instead hold the initial $\bz(\kappa)$-$\bD_1(r_1)$-$\bD_{2}(r_2)$-$\bl(d)$ structure fixed, which ensures that the rank of $\bL$ cannot be altered, and only train the second half using a standard $\ell_2$ loss.  In doing so we are able to obtain an $\bL$ matrix, extracted from the middle layer, that is both (a) \emph{not} well-represented by a low-rank approximation, and (b) \emph{does} lie on a \emph{known} low-dimensional non-linear manifold.  And given that essentially zero reconstruction error is in fact achievable (up to the expected small ripples introduced by stochastic gradient descent or a similar surrogate), the learned decoder from this process implicitly serves as the ground-truth encoder underlying the data structure.  Hence any VAE that includes $\bl(d)$-$\bE_1(r_2)$-$\bE_{2}(r_1)$-$\hat{\bz}(\kappa)$ within its encoder mean network capacity, as well as  $\bz(\kappa)$-$\bD_1(r_1)$-$\bD_{2}(r_2)$-$\hat{\bl}(d)$ within its decoder mean network capacity, will at least in principle have the capability of zero reconstruction error as well.

Finally, once $\bL$ has been created in this manner, we then generate the noisy data matrix $\bX$ by randomly corrupting $100 \cdot \nu \%$ of the entries, replacing the original value with samples from a standardized Gaussian distribution.  In doing so, the original `signal' component from $\bL$ is completely swamped out at these locations.

\vspace*{0.5cm}
\textbf{Experimental Design:}  Given a data matrix $\bX$ as generated above, we test the relative performance of four competing models:
\begin{enumerate}

\item \underline{\emph{VAE}}: We form a VAE architecture with the cascaded encoder/decoder mean networks $\bmu_x\left( \bmu_z\left[\bx \right] \right)$ assembled as $\bx(100)$-$\bE_1(2000)$-$\bE_2(1000)$-$\bmu_z(50)$-$\bD_1(1000)$-$\bD_{2}(2000)$-$\bmu_x(100)$. This mirrors the high-level structure used to generate the outlier-free data, and ultimately will ensure that the ground-truth manifold is included within the network parameterization.  Consistent with the design in~\cite{Kingma2014}, a diagonal encoder covariance $\bSigma_z$ is produced by sharing just the first two mean network layers.  An exponential layer is also appended at the output to produce non-negative values.  For consistency with AE models, the decoder covariance $\bSigma_x$ is addressed separately via a special process described below.

\item \underline{\emph{AE}-$\ell_2$}:  We begin with the VAE model from above and fix $\bSigma_z = {\bf 0}$.  This reduces the KL regularization term from (\ref{eq:KL_term}) to simply $\| \bmu_z \|_2^2$.  If no other changes are included, then the scaling ambiguity between $\bmu_z$ and decoder layer $\bD_1$ is such that $\bmu_z$ can be made arbitrarily small without any loss of generality, rendering any beneficial regularization effect from $\| \bmu_z \|_2^2$ completely moot as discussed at the beginning of Section \ref{sec:ppca}.  Therefore we add a standard weight decay term to the AE-$\ell_2$ network parameters $\{\btheta,\bphi\}$ to ameliorate this scaling ambiguity, which is tantamount to including an additional penalty factor $C_1 \| \{\btheta,\bphi\} \|_2^2$.  We also balance $\| \bmu_z \|_2^2$ with a second tuning parameter $C_2$, i.e., $C_2\| \bmu_z \|_2^2$.  For the experiments in this section, we choose $C_1 = 0.0005$, a typical default value for weight decay, and then tune $C_2$ for optimal performance.\footnote{For direct comparison, we include the same weight decay factor $C_1 \| \{\btheta,\bphi\} \|_2^2$ with the VAE model even though there is no equivalent issue with scaling ambiguity.  In fact, this can be viewed as an advantage of the VAE regularization mechanism, in that it directly prevents large decoder weights from compensating for arbitrarily small values of $\bmu_z$, killing regularization effects.  This is because there exists a key dependence between the weights from $\bD_1$ and the covariance $\bSigma_z$ such that any large weights that would accommodate pushing $\bmu_z$ towards zero would equally \emph{amplify} the random additive noise coming from the stochastic encoder model, nullifying any benefit to the overall cost.}

\vspace*{0.1cm}
Note also that once $\bSigma_z = {\bf 0}$, at every sample $\bSigma_x$ can be solved for in closed form as $\left[\bSigma^{(i)}_x\right]_{jj} = \left(x_j^{(i)} - \mu_{x_j}^{(i)} \right)^2$ for $j = 1,\ldots,d$ assuming sufficient capacity per Corollary \ref{cor:AE_behavior}.  We then plug this value into the AE-$\ell_2$ cost, effectively optimizing $\bSigma^{(i)}_x$ out of the model altogether making it entirely deterministic.  For direct comparison, we apply the same procedure to the VAE from above, which can be interpreted as efficiently modeling the infinite capacity limit for $\bSigma_x$ (i.e., even with infinite capacity in $\bSigma_x$, the VAE model could do no better than this).

    \item \underline{\emph{AE}-$\ell_1$}: To explicitly encourage sparse latent representations, which could potentially be helpful in learning the correct manifold dimension, we begin with the AE-$\ell_2$ model from above and replace $\| \bmu_z \|_2^2$ with the $\ell_1$ norm $\| \bmu_z \|_1$, a well-known sparsity-promoting penalty function \cite{Donoho03}.  The corresponding parameter $C_2$ is likewise independently tuned for optimal performance.

 \item \underline{RPCA}:  As an additional baseline, we also apply the convex RPCA formulation from~(\ref{eq:pcp}) to the same corrupted data.  This model is implemented via an augmented Lagrangian method using code from \cite{lin2010augmented}.

\end{enumerate}

For the VAE, AE-$\ell_2$, and AE-$\ell_1$ networks, all model weights were randomly initialized so as not to copy any information from the ground-truth template.  Training was conducted over 200 epochs using the Adam optimization technique \cite{kingma2014adam} with a learning rate of 0.0001 and a batch size of 100.  We chose $n = 10^6$ training samples for each separate experiment, across which we varied the manifold dimension from $\kappa = 2,4,\ldots,20$ while the outlier ratio ranged as $\nu = 0.05, 0.10, \ldots, 0.50$.  For each pair of experimental conditions, we train/run all four models and measure performance recovering the true $\bL$ as quantified by the normalized MSE metric
\begin{equation}
\mbox{NMSE} ~~ \triangleq ~~ \| \bL - \hat{\bL} \|_{\mathcal{F}}^2/\| \bL \|_{\mathcal{F}}^2.
\end{equation}
Note that although in practice we will not generally know the true manifold dimension $\kappa$ in advance, because we choose $\mbox{dim}[\bmu_z] = 50 > \kappa$ when constructing encoder networks for all experiments, perfect reconstruction is still theoretically possible by any of the VAE or AE models provided that outlier contributions can be successfully mitigated.

% ***

%We compare the VAE against the analogous AE as described above, and including the ground-truth manifold within the parameterization of the cascaded encoder/decoder mean networks $\bmu_x\left( \bmu_z\left[\bx \right] \right)$ for all models.  Therefore perfect reconstruction is theoretically possible by either approach provided that outlier contributions can be mitigated.

\begin{figure}[t!]
\begin{center}
  \subfigure[]{
    \includegraphics[width=0.3\textwidth]{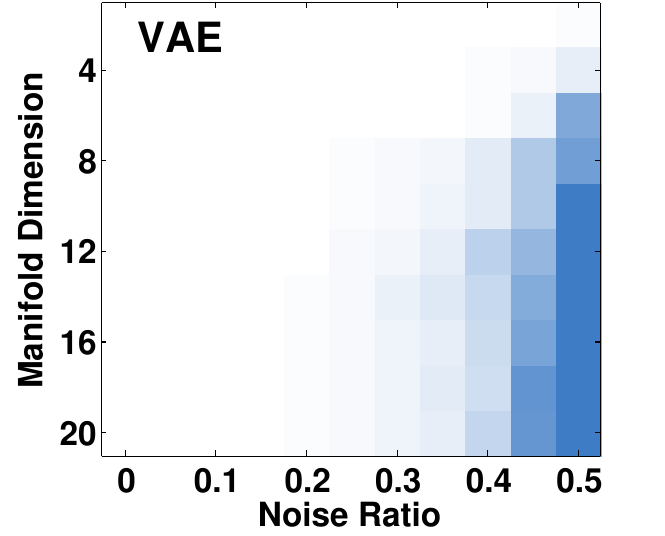}
    \label{fig:phase_transition_vae}}
  \subfigure[]{
    \includegraphics[width=0.3\textwidth]{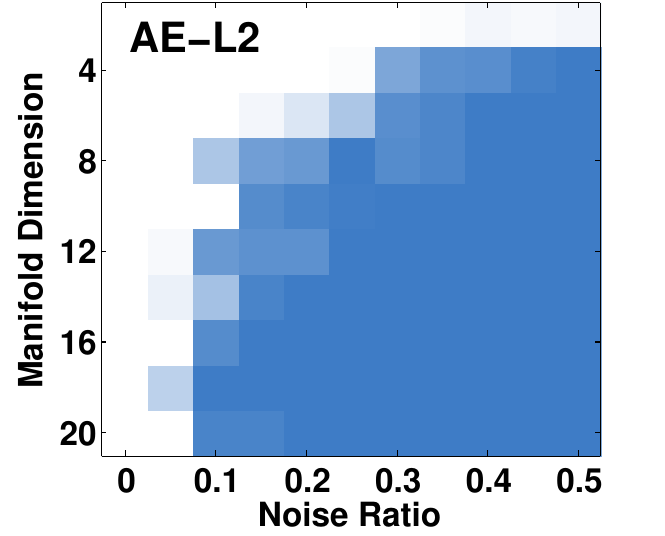}
    \label{fig:phase_transition_aeL2}} \\
  \subfigure[]{
    \includegraphics[width=0.3\textwidth]{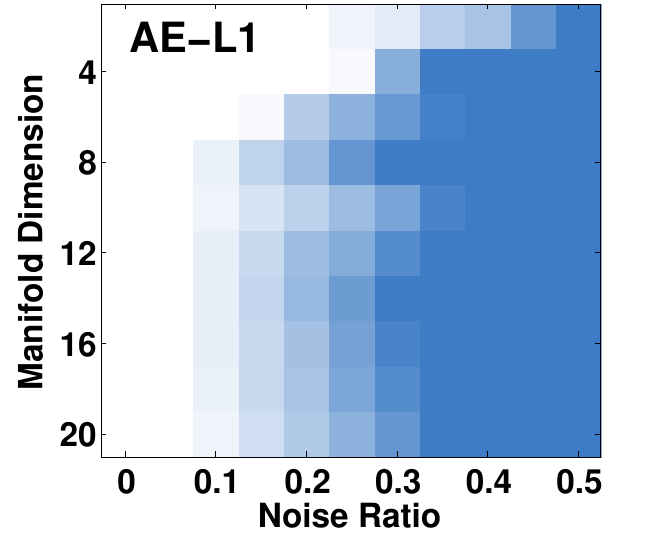}
    \label{fig:phase_transition_aeL1}}
  \subfigure[]{
    \includegraphics[width=0.3\textwidth]{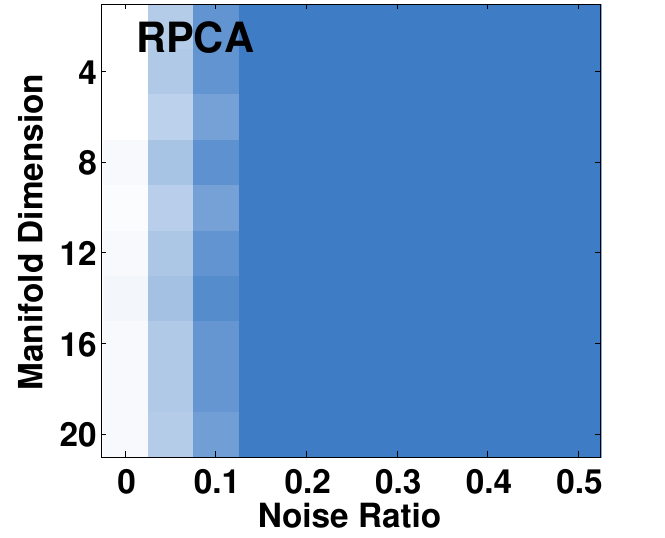}
    \label{fig:phase_transition_rpca}}
\end{center}
\caption{Results recovering synthesized low-dimensional manifolds across different outlier ratios ($x$-axis) and manifold dimensions ($y$-axis) for (\emph{a}) the VAE, (\emph{b}) the AE-$\ell_2$, (\emph{c}) the AE-$\ell_1$, and (\emph{d}) RPCA.  In all cases, white color indicates normalized MSE near $0.0$, while dark blue represents $1.0$ or failure.  The VAE is dramatically superior to each alternative, supporting Hypothesis (i).  Additionally, it is crucial to note here that the AE and RPCA solutions perform poorly for quite different reasons.  Not surprisingly, convex RPCA fails because it cannot accurately capture the underlying nonlinear manifold using a linear subspace inlier model.  In contrast, both AE-$\ell_1$ and AE-$\ell_2$ have the \emph{exact same} inlier model capacity as the VAE and can in principle represent uncorrupted points perfectly; however, they have inferior agency for pruning superfluous latent dimensions, discarding outliers, or generally avoiding bad locally-optimal solutions.}
\label{fig:phase_transition}
\end{figure}

\vspace*{0.5cm}
\textbf{Results:}  Figure \ref{fig:phase_transition} displays the results estimating $\bL$, where the VAE outperforms RPCA and the AE models by a wide margin. Perhaps most notably, the VAE performance dominates both AE-$\ell_1$ and AE-$\ell_2$, supporting our theory that the smoothing effect of integrating over $\bSigma_z$ has immense practical value in avoiding bad minimizing solutions through its unique form of differential regularization.  In fact, the AE-$\ell_2$ objective is identical to the VAE once $\bSigma_z = {\bf 0}$, at least up to the constant $C_2$ applied to $\| \bmu_z \|_2^2$ which is only tuned to benefit the former while remaining fixed for the latter.\footnote{If $C_2 = 1$, the default value as produced by the VAE KL term, the AE-$\ell_2$ performance is much worse (not shown) than when using the tuned value of $C_2 = 10^3$ as was adopted in producing Figure \ref{fig:phase_transition}.  In contrast, the VAE requires no such tuning at all, with the default $C_2 =1$ producing the results shown.}  So this smoothing effect is essentially the \emph{only} difference between the VAE and AE-$\ell_2$ models, and therefore, Figures \ref{fig:phase_transition_vae} and \ref{fig:phase_transition_aeL2} truly isolate the benefits of the VAE in this regard.

To summarize then, by design all VAE and AE network structures are equivalent in terms of their predictive capacity, but only the VAE is able to capitalize on the regularizing effect of $\bSigma_z$ to actually reach a good solution in challenging conditions.  Moreover, this is even possible without the hassle of tuning tedious hyperparameters to balance regularization effects as required by AE-$\ell_1$ and AE-$\ell_2$ models.\footnote{Of course we admittedly have not exhaustively ruled out the potential existence of some alternative regularizer capable of outperforming the VAE when carefully tuned to appropriate conditions; however, it is still nonetheless impressive that the VAE can naturally perform so well without such tuning on a task that it was not even originally motivated for.}  This confirms Hypothesis (i) and suggests that VAEs are a viable candidate for replacing existing RPCA algorithms \cite{Candes11,Ding11,oh2013partial,wipf2012} in regimes where a single linear subspace is an inadequate signal representation.  And we stress that, prior to the analysis herein, it was not at all apparent that a VAE could so dramatically outperform comparable AE models on this type of deterministic outlier removal task.

%We emphasize that this is \emph{not} a job description for which the VAE was originally motivated, but a useful hidden talent nonetheless.

%In contrast, we found that in many cases the AE had a lower MSE with respect to the noisy matrix $\bX$ than the clean $\bL$, a sign that it was stuck at a useless extrema which overfit the data.

Note also that perfect reconstruction, as consistently exhibited by the VAE in Figure \ref{fig:phase_transition_vae}, does \emph{not} actually require learning the correct generative model \emph{within} the estimated manifold.  Rather it only requires that as $\bSigma_z \rightarrow {\bf 0}$ selectively along appropriate dimensions (consistent with Hypothesis (iii) as will be discussed in Section \ref{sec:hypothesis_3_tests}), the encoder and decoder mean networks project onto the correct manifold while ignoring outliers.  Hence although random samples $\bz$ will likely lie on the true manifold when passed through the decoder network, they need not be perfectly distributed according to the full generative process unless sufficient additional capacity exists beyond that needed to represent the manifold itself.  We are not aware of this distinction being discussed in previous works, where VAE and related models are typically evaluated by either the overall quality of their generated samples \cite{dosovitskiy2016generating,larsen2015autoencoding,oord2016pixel}, or by the value of the likelihood bound \cite{Kingma2014,kingma2016improved,burda2015importance}.\footnote{Learning the correct distribution within the manifold, as required for full recovery of the entire generative process and the production of realistic samples, is a topic largely orthogonal to the analysis presented herein.  Still, to at least partially address these important issues, we have recently derived a VAE usage regime whereby, at least in principle, provable recovery within a manifold itself is possible even in situations where $\bSigma_z$ tends towards zero \cite{bin2019iclr}.}

%However, while in doing so the VAE may capture the manifold structure, when attempting to generate new samples they may not be distributed within the manifold in the same way as the training set.  This is because, as alluded to previously, accurate manifold recovery can be viewed as a necessary but not sufficient condition for actually reproducing the correct generative model.  For the latter, the VAE must also learn the correct distribution within this manifold, a topic largely orthogonal to the analysis presented in this work.  To at least partially address these important issues, we have recently derived relatively broad conditions whereby provable recovery within a manifold itself is possible; however, we defer presentation of this topic to future work.

% ***

Before proceeding to the next set of experiments, we address a tangential issue related to the RPCA performance as exhibited in Figure \ref{fig:phase_transition_rpca}.  When the outlier ratio is zero, RPCA can recover the ground-truth by simply defaulting to a full-rank inlier model without actually learning anything about the true manifold itself (the VAE and AE models do not have this luxury since they are forced to represent $\bL$ using at most $\mbox{dim}[\bmu_z] = 50 < \mbox{rank}[\bL] = 100$ dimensions by design).  In contrast, as the outlier ratio increases, it becomes increasingly difficult for RPCA to find any linear subspace representation that is both sufficiently high dimensional to include the majority of the inlier variance along the manifold while simultaneously excluding the outlier contributions.  This explains the steep drop-off in performance moving from left to right within Figure \ref{fig:phase_transition_rpca}.  But there is noticeably no change in RPCA performance as we move from top to bottom in the same plot.  This is because the clean data $\bL$ is full-rank regardless of the manifold dimension $\kappa$, and so any linear subspace approximation is more or less equally bad across all $\kappa$.

\subsection{Hypothesis (ii) Evaluation Using Ground-Truth Manifolds and MNIST Data}
\label{exp:evaluation_ii}

\textbf{Synthetic Data Example:}  To evaluate Hypothesis (ii), we train analogous AE and VAE models as the number of decoder and encoder hidden layers vary, in each case with ground-truth available per the procedure described above.  To generate each observed data point $\bx^{(i)}$, we sample $\bz^{(i)}$ from a 20-dimensional standard Gaussian distribution and pass it through a neural network structured as $\bz(20)$-$\bD_1(200)$-$\bD_2(200)$-$\bx(400)$, again with ReLU activations.  We then train VAE models of variable depth,  with concatenated mean networks $\bmu_x\left( \bmu_z\left[\bx \right] \right)$ designed as $\bx(400)$-$\bE_1(200)$-$...$-$\bE_{N_e}(200)$-$\bmu_z(30)$-$\bD_1(200)$-$...$-$\bD_{N_d}(200)$-$\bmu_x(400)$, where
$N_e$ and $N_d$ represent the number of hidden layers in the encoder and decoder respectively.  The corresponding covariances are modeled as in Section \ref{exp:evaluation_i}, and likewise, the training protocol is unchanged.  Note also that $\mbox{dim}[\bmu_z] = 30$ is considerably larger than the ground-truth dimension of 20.

The first layer of the decoder mean network (before the nonlinearity) can be expressed as
\begin{equation}
\bh_1 = \bW_1 \bz+\bb_1,
\end{equation}
which in isolation is equivalent to the affine decoder mean model.  If the VAE has the ability to find the true underlying manifold dimension, then the number of nonzero columns in $\bW_1$ should be $20$, indicating that $30-20=10$ dimensions of $\bz$ are actually useless for any subsequent representation, i.e., we can estimate the intrinsic dimension of the latent code by counting the number of nonzero columns in $\bW_1$, exactly analogous to the affine case.  Of course in practice it is unlikely that a column of $\bW_1$ converges all the way to exactly ${\bf 0}$ via any stochastic optimization method. Therefore we define a simple threshold as $\mbox{thr} = 0.05\times\max_{j=1}^{\kappa}||\bw_{\cdot j}||_2$. If $||\bw_{\cdot j}||_2< \mbox{thr}$, we regard it as a zero column.  But this heuristic notwithstanding, the partition between zero and non-zero columns is generally quite obvious as will be illustrated later.

Table~\ref{tab:non_zero} reports the estimated number of non-zero columns in $\bW_1$ as $N_e$ and $N_d$ are varied, where we have run $10$ trials for every pairing and averaged the results. When there is no hidden layer in the decoder (i.e., $N_d = 0$), which implies that the decoder mean is affine, all the columns are nonzero since the network is overly-simplistic and all degrees of freedom are being utilized to compensate. However, once we increase the depth, especially of the decoder within which $\bW_1$ actually resides, the number of nonzero columns of $\bW_1$ tends to exactly $20$, which is the correct ground-truth manifold dimension by design, directly supporting Hypothesis (ii).  Similar conclusions can be drawn from models of different sizes and configurations as well (not shown).  In contrast, we did not find a corresponding AE model with this capability.  % Additional experiments with MNIST data in the supplementary file likewise confirm these conclusions.

\begin{table}
\centering
\begin{tabular}{c|cccc}
\hline
 & $N_d=0$ & $N_d=1$ & $N_d=2$ & $N_d=3$ \\
\hline
$N_e=0$ & 30.0 & 21.1 & 21.0 & 20.0 \\
$N_e=1$ & 30.0 & 21.0 & 20.0 & 20.0 \\
$N_e=2$ & 30.0 & 21.0 & 20.0 & 20.0 \\
$N_e=3$ & 30.0 & 20.4 & 20.0 & 20.0 \\
\hline
\end{tabular}
\vspace{0.1cm}
\caption{Number of nonzero columns in the VAE decoder mean first-layer weights $\bW_1$ learned using different encoder and decoder depths applied to data with a ground-truth latent dimension of 20.  Provided that the VAE model is sufficiently complex, the correct estimate is automatically obtained. We have not found an analogous AE model with similar capability.}
\label{tab:non_zero}
\end{table}

Note that prior work has loosely suggested that the KL regularizer indigenous to VAEs could potentially mute the impact of superfluous latent dimensions as part of the model optimization process~\cite{burda2015importance,sonderby2016train}.  However, there has been no theoretical or empirical demonstration of why this should happen, nor any rigorous explanation of a precise pruning mechanism built into the aggregate VAE cost function itself.  And as mentioned previously, the KL term is characterized by an $\ell_2$ norm penalty on $\bmu_z$ (see (\ref{eq:KL_term})), which we would normally expect to promote low-energy latent representations with mostly small, but nonzero values \cite{Chen99}, the exact \emph{opposite} of any sparsity-promotion or pruning agency.  But of course if columns of $\bW_1$ are set to zero, then no information about $\bz$ can pass through these dimensions to the hidden layers of the decoder.  Therefore the KL term can now be minimized in isolation along these dimensions with the corresponding elements of $\bmu_z$ set to exactly zero. Hence it is only the counterintuitive co-mingling of \emph{all} energy terms that leads to this desirable VAE pruning effect as we have meticulously characterized.

\begin{figure}[t!]
\begin{center}
  \subfigure[]{
    \includegraphics[width=0.5\linewidth]{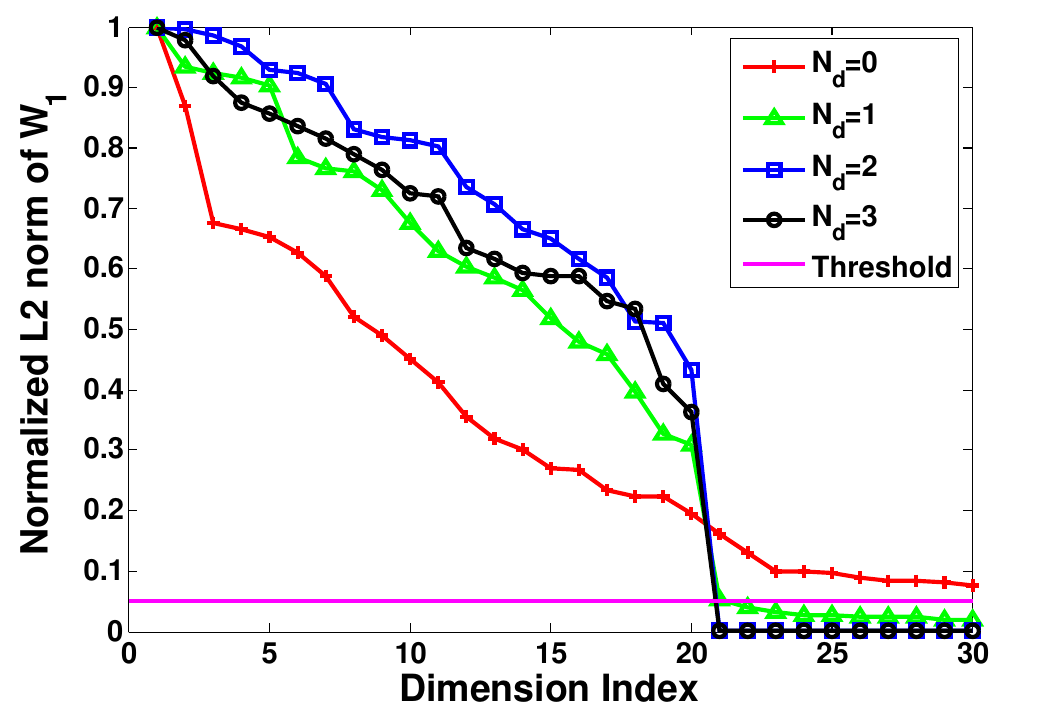}
    \label{fig:wo_decoder}}
  \subfigure[]{
    \includegraphics[width=0.4\linewidth]{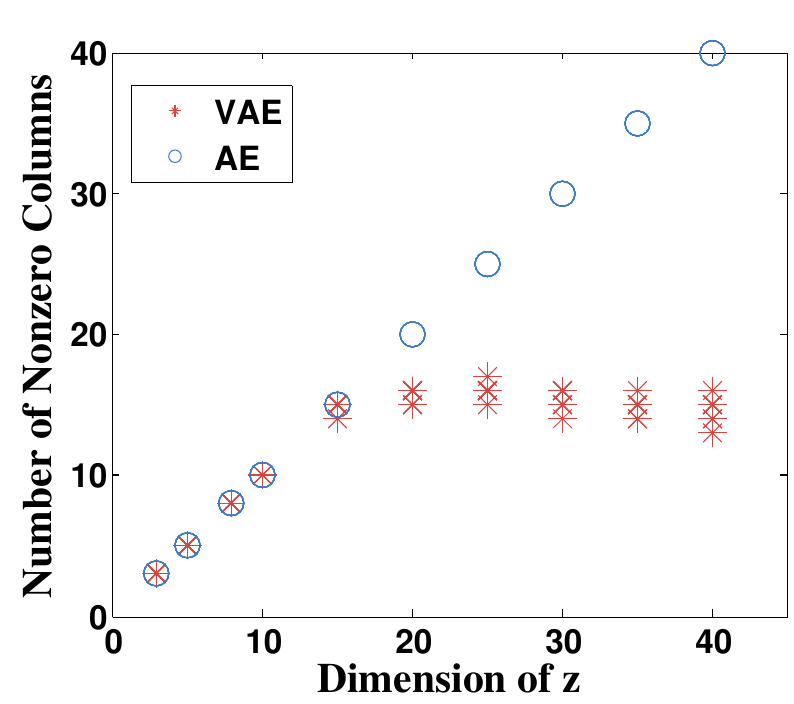}
    \label{fig:non_zero}}
\end{center}
\caption{(\emph{a}) Validation of thresholding heuristic for determining nonzero columns in $\bW_1$.  With $N_e = 3$ and the settings from Table \ref{tab:non_zero}, the sorted column norms of $\bW_1$ are plotted.  Clearly for $N_d \in \{2,3\}$ the gap between zero and nonzero values is extremely clear and any reasonable thresholding heuristic will suffice.  (\emph{b}) Number of nonzero columns in the decoder mean first-layer weights $\bW_1$ as the latent dimension $\kappa$ is varied for both AE and VAE models trained on MNIST data. Only the VAE automatically self-regularizes when $\kappa$ becomes sufficiently large (here at $\kappa\approx 15$), consistent with Hypothesis (ii).}
\end{figure}

Finally, Figure \ref{fig:wo_decoder} provides validation for our heuristic criterion for classifying columns of $\bW_1$ as zero or not.  Under the same experimental conditions as were used for creating Table \ref{tab:non_zero}, we plot the sorted column norms of $\bW_1$ for the cases where $N_e = 3$ and $N_d \in \{0,1,2,3\}$.  Especially when $N_d \in \{2,3\}$, meaning the model is of (or nearly of) sufficient capacity, zero and nonzero values are easily distinguishable and any reasonable thresholding heuristic would be adequate. Likewise for $N_d=0$ it is clear that all values are significantly distant from zero.  In contrast, when $N_d = 1$ (green curve) it is admittedly more subjective whether or not the smallest 9 or 10 elements should be classified as zero.  Regardless, the overall trend is unequivocal, with any heuristic threshold only influencing the $N_d = 1$ boundary case.

\vspace*{0.5cm}
\textbf{MNIST Example:} To further verify Hypothesis (ii), we train VAE models on the MNIST dataset of handwritten digit images \cite{Lecun1998} as $\kappa$ is varied. We use all $n=70000$ samples, each of size $28\times 28$.  We structure  4-layer cascaded VAE mean networks $\bmu_x\left( \bmu_z\left[\bx \right] \right)$ as $\bx(d)$-$\bE_1(1000)$-$\bE_2(500)$-$\bE_3(250)$-$\bmu_z(\kappa)$-$\bD_1(250)$-$\bD_{2}(500)$-$\bD_{3}(1000)$-$\bmu_x(d)$, where $d = 28\times 28 = 784$ and ReLU activations are used.  Covariances and training protocols are handled as before.  We draw values of $\kappa$ from $\{3, 5, 8, 10, 15, 20, 25, 30, 35, 40 \}$.

Figure~\ref{fig:non_zero} displays the number of nonzero columns in $\bW_1$ produced by each $\kappa$-dependent model, again across 10 trials.  We observe that when $\kappa > 15$, the number of nonzero columns plateaus for the VAE consistent with Hypothesis (ii).  Of course unlike the synthetic case, we no longer have access to ground truth for determining what the optimal manifold dimension should be.

We also applied an analogous AE model trained with $C_2 = 0$, i.e., a standard AE with no additional regularization penalty added.  Not surprisingly, the number of nonzero columns in $\bW_1$ is always equal to $\kappa$ since there is no equivalent agency for column-wise pruning as implicitly instilled by the VAE.   Note that tuning $C_2$ with either $\ell_1$- or $\ell_2$-norm penalties is of course always possible; however, the optimal value can be $\kappa$-dependent making subsequent results less interpretable.  Moreover, in general we have not found a setting whereby the penalties lead to correct latent dimensionality estimation in situations where the ground-truth is known.

\subsection{Hypothesis (iii) Evaluation Using Covariance Statistics from Corrupted Manifold Recovery Task} \label{sec:hypothesis_3_tests}
\begin{figure*}
\begin{center}
\includegraphics[width=1\linewidth]{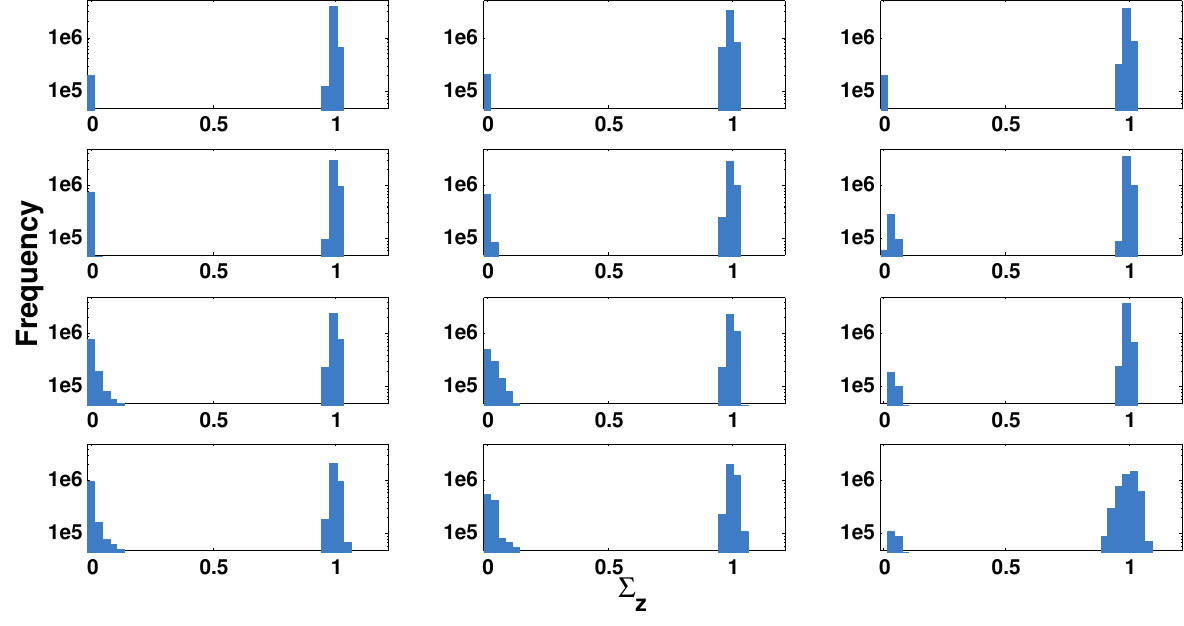}
\end{center}
\caption{Log-scale histograms of $\left\{ \bSigma_z^{(i)} \right\}_{i=1}^n$ diagonal elements as outlier ratios and manifold dimensions are varying for the corrupted manifold recovery experiment corresponding with Figure~\ref{fig:phase_transition}.  The three columns represent outlier ratios of $\nu \in \{0.0, 0.25, 0.50\}$ from left to right. The four rows represent manifold dimensions of $\kappa \in \{2, 8, 14, 20\}$ from top to bottom.  All plots demonstrate the predicted clustering of variance values around either zero or one.  Likewise, the relative sizes of these clusters, including observed changes across experimental conditions, conforms with our theoretical predictions (see detailed description in Section \ref{sec:hypothesis_3_tests}).}
\label{phase_transition_var_z}
\end{figure*}
%When the noise ratio and the manifold dimension are both very low (easy case), almost all the $\bSigma_z^{(i)}$ diagonal elements become very close to $0$, which means the network has nearly perfectly fit the inliers and rejected the outliers, consistent with theory.  As the noise ratio and manifold dimension increase (harder case), some elements of $\bSigma_z^{(i)}$ gradually drift away from $0$ and becomes very close to $1$; in this regime the VAE has not perfectly modeled all of the data, and the KL penalty pushes these covariances towards the value minimizing the KL term.

% ***********************

If some columns of $\bW_1$ tend to zero as we have argued both empirically and theoretically, then the corresponding diagonal elements of $\bSigma_z$, like $\bmu_z$, can no longer influence the decoder.  And with only the lingering KL term to offer guidance, along these coordinates the optimal variance will then equal one by virtue of (\ref{eq:sigma_z_identity}).  But for nonzero columns of $\bW_1$, the behavior of $\bSigma_z$ is much more counter-intuitive.  Despite the $-\log|\bSigma_z|$ factor from the KL divergence that contributes an unbounded cost as any $\left[ \bSigma_z \right]_{jj} \rightarrow 0$, we nonetheless have proven for the affine decoder mean case a natural tendency of the VAE  to push these variance values arbitrarily close to zero when approaching globally optimal solutions, at least along latent dimensions required for representing inlier points lying on ground-truth manifolds (i.e, dimensions where $\bW_1$ is nonzero).

%To empirically verify the natural tendency of the VAE to push the covariances $\bSigma_z^{(i)} = \bSigma_z(\bx^{(i)} ; \bphi)$ to zero when approaching optimal solutions along latent dimensions required for capturing ground-truth manifolds, we slightly modify the testing conditions from Section~\ref{exp:evaluation_i}.  Instead of choosing $\mbox{dim}[\bmu_z] = 50$, we set this value to the ground-truth manifold dimension.  In this way, we expect that perfect recovery should require all encoder covariance elements to be pushed towards zero.

%We now empirically verify that this same effect is inherited by general VAE models with more sophisticated, nonlinear decoder mean networks.  For this purpose, we created histograms of all diagonal elements of $\left\{ \bSigma_z^{(i)} \right\}_{i=1}^n$ obtained from the experiments described in Section~\ref{exp:evaluation_i} where the noise ratios and manifold dimensions vary.

We now empirically verify that this same effect is inherited by general VAE models with more sophisticated, nonlinear decoder mean networks.  For this purpose, we created histograms of all diagonal elements of $\left\{ \bSigma_z^{(i)} \right\}_{i=1}^n$ obtained from the experiments described in Section~\ref{exp:evaluation_i} where the outlier ratios and manifold dimensions vary.  The results are plotted in Figure~\ref{phase_transition_var_z} for all pairs of outlier ratios $\nu \in \{0.0,0.25,0.50 \}$ (columns) and ground-truth manifold dimensions $\kappa \in \{2, 8, 14, 20\}$ (rows).  These results directly conform with our theoretical predictions per the following explanations.

First, consider the upper-left panel displaying the simplest case from an estimation standpoint, since $\nu = 0.0$ (no outliers) and $\kappa = 2$ (very low-dimensional manifold).  Here we observe a clear partitioning between elements of $\left\{ \bSigma_z^{(i)} \right\}_{i=1}^n$ going to either zero or one.  Moreover, given that $\mbox{dim}[\bmu_z] = 50$ while the ground-truth involves $\kappa = 2$, 48 out of 50 dimensions are actually unnecessary.  Hence we should expect that only about $4\%$ of variance values should concentrate around zero, with the remainder forced towards one.  In fact, this is precisely the general partitioning we observe (note the log scaling of the y-axis).  Additionally, if we examine the other panels in the left-most column of Figure~\ref{phase_transition_var_z}, we notice that as the ground-truth $\kappa$ increases, the percentage of variance values shifts from one to zero roughly proportional to $\kappa/50$.  In other words, as more dimensions are required to represent the more challenging, higher-dimensional manifolds, more diagonal elements of each $\bSigma_z^{(i)}$ are pushed towards zero to enforce accurate reconstructions.

Next, we observe that in the top row of Figure~\ref{phase_transition_var_z}, each of the three panels are more or less the same, indicating that the inclusion of outliers has not disrupted the VAE's ability to model the ground-truth manifold.  In contrast, the bottom row presents a somewhat different story.  Given the more challenging conditions with a much higher dimensional ground-truth manifold ($\kappa = 20$), the inclusion of additional outliers (as we move from left to right) shifts more variance elements from zero to one.  This implies that the VAE, when confronted with both a higher-dimensional manifold and severe outliers (bottom-right panel), is settling on a relatively lower-dimensional approximation.  This behavior is reasonable in the sense that accurately estimating a complex manifold via any method becomes problematic when $50\%$ of the data is corrupted, and a low-dimensional approximation is all that is feasible to avoid simply fitting all the outliers.  In this situation some manifold dimensions of lesser importance can be viewed as expendable, and consequently we will likely have additional elements of $\left\{ \bSigma_z^{(i)} \right\}_{i=1}^n$ tending to one.

To further elucidate this phenomena, we include one additional supporting visualization involving the special case where the ground-truth manifold dimension equals 10 while the outlier ratio $\nu$ varies.  However, we slightly modify the testing conditions from Section~\ref{exp:evaluation_i}.  Instead of choosing $\mbox{dim}[\bmu_z] = 50$, we set this value to the ground-truth value $\kappa = 10$.  In this constrained setting, we expect that perfect recovery should require \emph{all} diagonal elements of $\bSigma_z$ to be pushed towards zero, since there are no longer any superfluous degrees of freedom.  Therefore, if any covariance elements tend to one, we have isolated the emergence of a low-dimensional approximation as presumably necessitated by increasing outlier levels.

\begin{figure*}
\begin{center}
\includegraphics[width=0.5\linewidth]{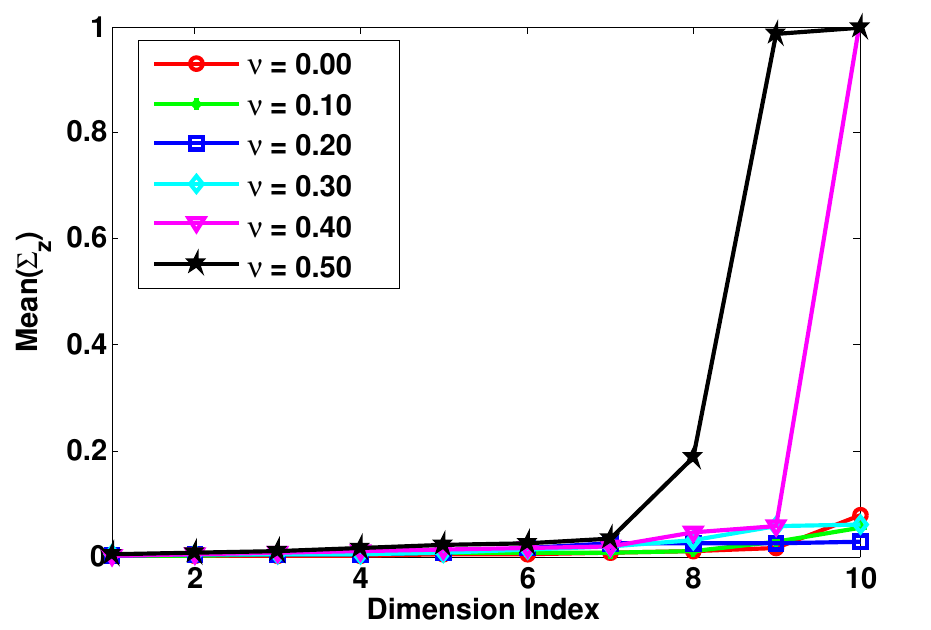}
\end{center}
\caption{Diagonal values of $\tfrac{1}{n}\sum_{i=1}^n \bSigma_z^{(i)}$ sorted in ascending order for a VAE model trained with ground-truth $\kappa = \mbox{dim}[\bmu_z] = 10$ on the recovery task from Section~\ref{exp:evaluation_i}.  When the outlier proportion is $\nu \leq 0.30$, the average variance is near zero across all latent dimensions.  However, for $\nu > 0.30$, some variance values are pushed towards one, indicating that the VAE is defaulting to a lower-dimensional approximation.}
\label{fig:mean_var_z_true}
\end{figure*}

Figure~\ref{fig:mean_var_z_true} displays the diagonal values of $\tfrac{1}{n} \sum_{i=1}^n \bSigma_z^{(i)}$ sorted in ascending order along the x-axis.  When $\nu \leq 0.30$, the average variance is near zero across all latent dimensions.  However, for $\nu > 0.30$, some variance values are pushed towards one, indicating that the VAE is defaulting to a lower-dimensional approximation.

% ***

To summarize then, the results of this section help to confirm a rather curious behavior of the VAE:  If $\bmu_x\left( \bmu_z\left[\bx \right] \right)$ is suitably parameterized to model inlier samples, and $\bSigma_x$ is sufficiently complex to model outlier locations, then elements of $\bSigma_z$ can be selectively pushed towards zero in the neighborhood of global minima. This involves overpowering the $-\log|\bSigma_z\|$ factor from the KL divergence that would otherwise seemingly prevent this from happening.  Moreover, in this degenerate regime, the VAE will exhibit deterministic behavior and can perfectly represent original clean training data samples via a low-dimensional manifold provided that the outlier level is not too high.

\section{Discussion}
\label{sec:discussion}
Although originally developed as a viable deep generative model or tractable bound on the data likelihood, in this work we have revealed certain properties and abilities of the VAE that are not obvious from first inspection.  For example, in addition to its putative role in driving diversity into the learned generative process, the latent covariance $\bSigma_z$ also serves as an important smoothing mechanism that aids in the robust recovery of corrupted samples, even if sometimes this requires exhibiting behavior (i.e., selective convergence towards zero) that may seem counterintuitive.  And although the VAE only adopts an $\ell_2$ norm penalty on $\bmu_z$ that in isolation should favor low energy solutions with all or mostly nonzero values, the latent mean estimator nonetheless tends to be highly sparse because of subtle, non-obvious interactions with other factors in the energy function such as the first-layer decoder mean network weights $\bW_1$.  Likewise, outliers can be estimated and completely removed via the action of $\bSigma_x$ despite no traditional, additive sparsity penalty applied across each data point.

In general, our results speak to many under-appreciated aspects of VAE behavior, have wide ranging practical consequences, and suggest novel usages beyond the original VAE design principles.  These include:
\begin{itemize}
\item The VAE can be applied to estimating deterministic nonlinear manifolds heavily corrupted with outliers.

\item The self-regularization effects of the VAE can largely handle excessive degrees of freedom when it comes to the latent representation $\bz$ as produced by the full encoder and processed by the first layer of the decoder mean network, as well as an arbitrarily-parameterized decoder covariance $\bSigma_x$.  Conversely, only excessive complexity specifically localized in higher decoder mean network layers can, at least in principle, lead to potential problems with overfitting.

\item The latent covariance $\bSigma_z$ can serve as an approximate bellwether for determining the true dimensionality of a manifold, provided that excessive outliers/corruptions do not lead to an under-estimate.  This is because typically near global solutions, we observe $[\bSigma_z]_{jj} \rightarrow 0$ for \emph{useful} dimensions, while for \emph{useless} dimensions we have shown that $[\bSigma_z]_{jj} \rightarrow 1$, a clear bifurcation.

\end{itemize}
Although the primary purpose of this paper is not to build a better generative model per se, we nevertheless hope that ideas introduced here will help to ensure that VAEs are not under or improperly utilized.  Additionally, in closing we should also mention that the focus herein has been almost entirely on the analysis of the VAE energy function itself, independent of the specifics of how this energy function might ultimately be optimized in practice.  But we believe the latter to be an equally-important, complementary topic, and further study is undeniably warranted.  For example, if an optimization trajectory is somehow lured astray by the Siren's song of a bad local minimum in the VAE energy landscape, then obviously many of our conclusions predicated on global optima will not necessarily still hold.

% \vspace*{4cm}

%==================================================
% Acknowledgements should go at the end, before appendices and references
\acks{Y. Wang is sponsored by the EPSRC Centre for Mathematical Imaging in Healthcare, University of Cambridge, EP/N014588/1. Y. Wang is also partially sponsored by Microsoft Research, Beijing.}


\newpage

%==================================================
% Appendix
\appendix
%========================================================

\section{Additional MNIST Dataset Experiment}

 Here we examine practical denoising of MNIST data corrupted with outliers using a VAE model. Outliers are added to MNIST handwritten digit data \cite{Lecun1998} by randomly replacing from $5\%$ to $50\%$ of the pixels with a value uniformly sampled from $[0,255]$ to create $\bX$.  We choose $\kappa = 30$ for the dimension of $\bz$ and apply the same VAE structure as applied to MNIST data in Section \ref{exp:evaluation_ii}.  The model is trained using both $\tau = 1$ and $\tau = 5$ latent samples $\{\bz^{(i,t)}\}_{t=1}^{\tau}$ for each $\bx^{(i)}$, observing that the latter, which more closely approximates the posterior, should perform significantly better.

% The encoder is the same as in Section \ref{exp:evaluation_ii}, \emph{i.e.}, a 4-layer ReLU network ($d$-$1000$-$500$-$250$-$\kappa$), where we choose $\kappa = 30$ for the dimension of $\bz$.

\begin{figure}
\begin{center}
  \subfigure[]{
    \includegraphics[width=0.45\linewidth]{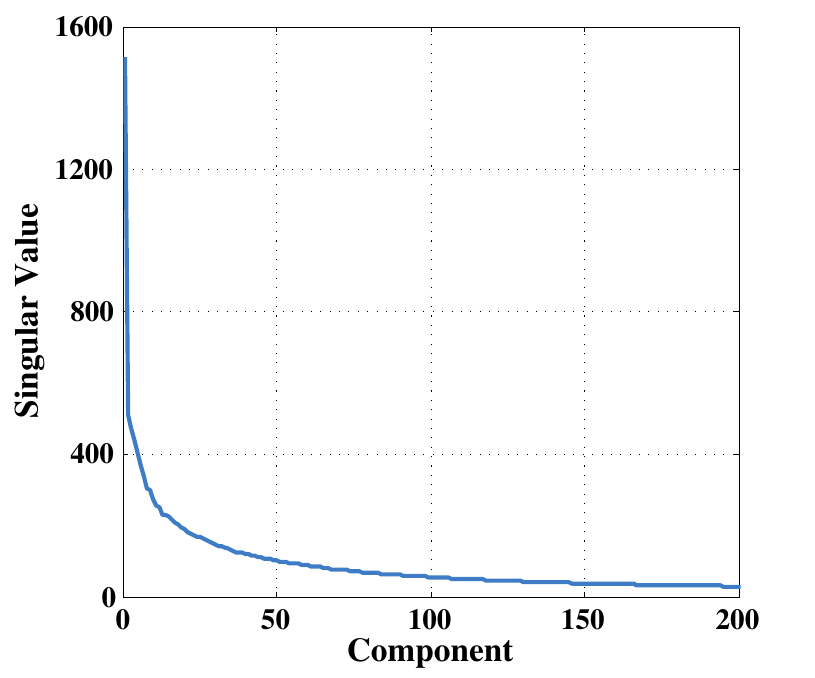}
    \label{fig:mnist_svd}
  }
  \subfigure[]{
    \includegraphics[width=0.45\linewidth]{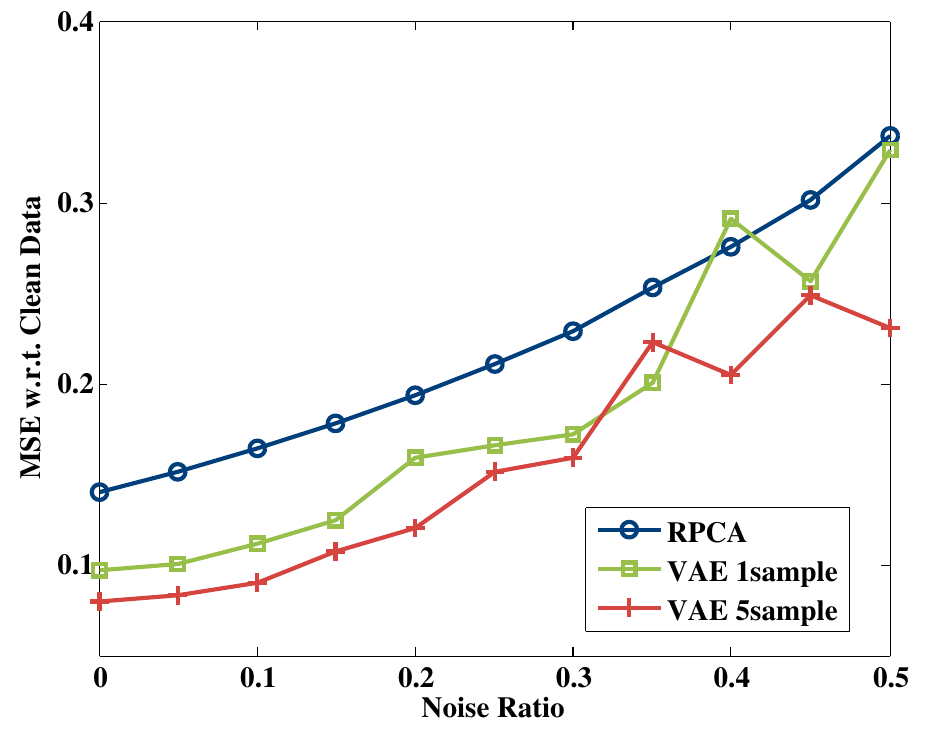}
    \label{fig:mnist_mse}
  }
\end{center}
\caption{(\emph{a}) Singular value spectrum of MNIST data revealing (approximately) low-rank structure. (\emph{b}) Normalized MSE recovering MNIST digits from corrupted samples.  The VAE is able to reduce the reconstruction error by better modeling more fine-grain details occupying the low end of the singular value spectrum.}
\label{fig:mnist}
\end{figure}

We compare the VAE against convex RPCA on the task of recovering the original, uncorrupted digits.  Note that RPCA is commonly used for unsupervised cleaning of this type of data \cite{Elhamifar13}, and MNIST is known to have significant low-rank structure \cite{lu2013correlation} as shown in Figure~\ref{fig:mnist_svd}.  Regardless, we observe in Figure~\ref{fig:mnist_mse} that the VAE performs significantly better in terms of normalized MSE by capturing additional manifold details that deviate from a purely low-rank representation.  Furthermore, we hypothesize that using extra latent samples (the $\tau = 5$ case) may work better on outlier removal tasks given the strong need for accurate smoothing of the VAE objective as described previously.

%Here we examine practical denoising of MNIST data corrupted with outliers using a VAE model.

%========================================================
\section{Proof of Lemma~\ref{lem:ppca}}
\label{sec:proof_lemma1}
Under the stated assumptions, the VAE cost can be simplified as
\begin{eqnarray} \label{eq:vae_cost_tot_affine_sup}
\calL(\btheta,\bphi) &  =  & \sum_i \left\{ \mathbb{E}_{q_{\tiny \bphi}\left(\bz|\bx^{(i)} \right)} \left[ \tfrac{1}{\lambda} \left\| \bx^{(i)} - \bW \bz   - \bb  \right\|_2^2  \right] + d \log \lambda  \right. \nonumber \\
& & +  ~~ \left.  \mbox{tr}\left[ \bSigma_z^{(i)} \right] - \log \left| \bSigma_z^{(i)}   \right| + \| \bmu_z^{(i)} \|_2^2  \right\} \nonumber \\
& = & \sum_i \left\{ \tfrac{1}{\lambda} \left\| \bx^{(i)} - \bW \bmu_z^{(i)}   - \bb  \right\|_2^2 + \tfrac{1}{\lambda} \mbox{tr}\left[\bSigma_z^{(i)} \bW^{\top} \bW  \right] + d \log \lambda  \right. \nonumber \\
& & +  ~~ \left.  \mbox{tr}\left[ \bSigma_z^{(i)} \right] - \log \left| \bSigma_z^{(i)}   \right| + \| \bmu_z^{(i)} \|_2^2  \right\},
\end{eqnarray}
where $\bmu_z^{(i)} \triangleq \bmu_z\left(\bx^{(i)}; \bphi \right)$ and $\bSigma_z^{(i)} \triangleq \bSigma_z\left(\bx^{(i)} ; \bphi \right)$.  Given that
\begin{equation}
\log\left| \bA\bA^{\top} \right| = \arg\inf_{\bGamma \succ {\bf 0}} \mbox{tr}\left[\bA\bA^{\top} \bGamma^{-1} \right] + \log\left| \bGamma \right|,
\end{equation}
when optimization is carried out over positive definite matrices $\bGamma$, minimization of (\ref{eq:vae_cost_tot_affine_sup}) with respect to $\bSigma_z^{(i)}$ leads to the revised objective
\begin{eqnarray} \label{eq:vae_cost_tot_affine2_sup}
\calL(\btheta,\bphi) &  \equiv  & \sum_i \left\{ \tfrac{1}{\lambda} \left\| \bx^{(i)} - \bW \bmu_z^{(i)}   - \bb  \right\|_2^2 + \log\left|\tfrac{1}{\lambda} \bW^{\top} \bW  + \bI \right| + d \log \lambda  + \| \bmu_z^{(i)} \|_2^2  \right\} \nonumber \\
&  =  & \sum_i \left\{ \tfrac{1}{\lambda} \left\| \bx^{(i)} - \bW \bmu_z^{(i)}   - \bb  \right\|_2^2 + \log\left|\bW \bW^{\top} + \lambda \bI  \right| +  \| \bmu_z^{(i)} \|_2^2  \right\},
\end{eqnarray}
ignoring constant terms.  This expression only requires that $\bSigma_z^{(i)} = \left[ \tfrac{1}{\lambda} \bW^{\top} \bW  + \bI \right]^{-1}$, or a constant parameterization, independent of $\bx^{(i)}$.  Similarly we can optimize over $\bmu_z^{(i)}$ in terms of the other variables.  This is just a ridge regression problem, with optimal solution
\begin{equation} \label{eq:mu_z_opt}
\bmu_z^{(i)} = \bW^{\top} \left( \lambda \bI + \bW\bW^{\top} \right)^{-1} \left(\bx^{(i)}-\bb \right),
\end{equation}
or a simple linear function of $\bx^{(i)}$.  Hence as long as the parameterization of both $\bmu_z^{(i)}$ and $\bSigma_z^{(i)}$ allows for arbitrary affine functions as stipulated in the lemma statement, these optimal solutions are feasible.  Plugging (\ref{eq:mu_z_opt}) into (\ref{eq:vae_cost_tot_affine2_sup}) and  applying some basic linear algebra, we arrive at
\begin{eqnarray} \label{eq:vae_cost_tot_affine3_sup}
\calL(\btheta,\bphi) &  \equiv  &  \sum_i \left( \bx^{(i)} - \bb \right)^{\top} \left( \bW \bW^{\top}  + \lambda \bI\right)^{-1}\left( \bx^{(i)} - \bb \right)  + n \log\left|\bW \bW^{\top} + \lambda \bI  \right|.
\end{eqnarray}

Finally, in the event that we enforce that $\bSigma_z^{(i)}$ be diagonal, (\ref{eq:vae_cost_tot_affine2_sup}) must be modified via
\begin{equation}
\bSigma_z^{(i)} = \left[ \tfrac{1}{\lambda} \mbox{diag}\left(\mbox{diag}\left[\bW^{\top} \bW \right]\right) + \bI \right]^{-1} = \sum_{j=1}^{\kappa} \log \left( \lambda + \| \bw_{\cdot j} \|_2^2 \right) - \kappa \log \lambda,
\end{equation}
where the $\mbox{diag}[\cdot]$ operator converts vectors to diagonal matrices, and a matrix to a vector formed from its diagonal (just as in the Matlab computing environment), leading to the stated result.

%========================================================
\section{Proof of Theorem \ref{thm:ppca_minima}}
\label{sec:proof_theorem1}

First, for part 1 on the theorem, given that $\bW \bR \bR^{\top} \bW^{\top} = \bW \bP \bP^{\top} \bW^{\top} = \bW \bW^{\top}$ for any rotation $\bR$ and permutation $\bP$, then obviously if $\bW^*$ is a minimum of (\ref{eq:ppca_cost}), $\bW^* \bR$ and $\bW^* \bP$ must also be.  Likewise, since $\sum_{j=1}^{\kappa} \log \left( \lambda + \| \bw_{\cdot j} \|_2^2 \right)$ is invariant to the order of the summation, then if $\bW^{**}$ is a minimum of (\ref{eq:ppca_cost_separable}), $\bW^{**} \bP$ must be as well.

We also have that
\begin{eqnarray} \label{eq:thm1_bound_sup}
\calL_{\tiny sep}(\bW^{**}, \bb, \lambda) & = &  \sum_i \bOmega^{(i)}(\bW^{**},\bb,\lambda \bI) +  n \left[ \sum_j \log \left(\lambda + \|\bw^{**}_{\cdot j} \|_2^2  \right) + (d-\kappa)\log \lambda \right] \nonumber \\
& = & \sum_i \bOmega^{(i)}(\bW^{**},\bb,\lambda \bI) +  n \left[ \sum_j \log \left(1 + \tfrac{1}{\lambda} \|\bw^{**}_{\cdot j} \|_2^2  \right) + d \log \lambda \right] \nonumber \\
& \geq & \sum_i \bOmega^{(i)}(\bW^{**},\bb,\lambda \bI) +  n\left[ \log \left| \tfrac{1}{\lambda} \left(\bW^{**}\right)^{\top} \bW^{**}  + \bI \right| + d \log \lambda \right] \nonumber \\
& = & \sum_i \bOmega^{(i)}(\bW^{**},\bb,\lambda \bI) +  n\log \left| \lambda \bI + \bW^{**} \bR \left(\bW^{**} \bR \right)^{\top}   \right| \nonumber \\
& \geq & \sum_i \bOmega^{(i)}(\bW^{*},\bb,\lambda \bI) +  n\log \left| \lambda \bI + \bW^* \bR \left(\bW^* \bR \right)^{\top}   \right|,
\end{eqnarray}
where the the second inequality follows from the fact that $\bW^*$ is an optimal solution to (\ref{eq:ppca_cost}).  The first inequality stems from Hadamard's inequality \cite{Garling07} applied to
\begin{equation}
\tfrac{1}{\lambda} \left(\bW^{**}\right)^{\top} \bW^{**}  + \bI = \bM^{\top} \bM
\end{equation}
for some square matrix $\bM$ of appropriate dimension.  This results in
\begin{equation}
\log \left| \tfrac{1}{\lambda} \left(\bW^{**}\right)^{\top} \bW^{**}  + \bI \right| = 2 \log\left|\bM \right| \leq 2\log \left( \prod_j \| \bm_{\cdot j} \|_2\right) = \sum_j \log \left(1 + \tfrac{1}{\lambda} \|\bw^{**}_{\cdot j} \|_2^2  \right),
\end{equation}
with equality iff $\bM^{\top} \bM$ is diagonal.  We can further manipulate the log-det term in (\ref{eq:thm1_bound_sup}) via
\begin{eqnarray}
n\log \left| \lambda \bI + \bW^* \bR \left(\bW^* \bR \right)^{\top}   \right| & = & \log\left|\tfrac{1}{\lambda} \left(\bW^* \bR \right)^{\top} \bW^* \bR  + \bI \right| + d \log \lambda \nonumber \\
& = & \log\left|\tfrac{1}{\lambda} \left(\bU\bLambda \bV^{\top} \bR \right)^{\top} \bU\bLambda \bV^{\top} \bR  + \bI \right| + d \log \lambda \nonumber \\
& = & \log\left|\tfrac{1}{\lambda} \bR^{\top} \bV \bLambda^2 \bV^{\top} \bR  + \bI \right| + d \log \lambda, \nonumber
\end{eqnarray}
where $\bU \bLambda \bV^{\top}$ is the SVD of $\bW^*$.  Now if we choose $\bR = \bV$ and define $\bar{\bW} \triangleq \bW \bV$, then $\Lambda_{jj} = \| \bar{\bw}_{\cdot j} \|_2^2$ and this expression further reduces via
\begin{eqnarray}
\log\left|\tfrac{1}{\lambda} \bR^{\top} \bV \bLambda^2 \bV^{\top} \bR  + \bI \right| + d \log \lambda & = & \sum_j \log \left(1 + \tfrac{1}{\lambda} \Lambda_{jj}  \right) + d \log \lambda \nonumber \\
& = & \sum_j \log \left(\lambda +  \| \bar{\bw}_{\cdot j} \|_2^2  \right) + (d-\kappa) \log \lambda.
\end{eqnarray}
Of course we cannot have
\begin{eqnarray}
&& \hspace*{-1.2cm} \sum_i \bOmega^{(i)}(\bW^{**},\bb,\lambda \bI) +  n \left[\sum_j \log \left(\lambda + \|\bw^{**}_{\cdot j} \|_2^2  \right) + (d-\kappa)\log \lambda \right]  \\
& & \hspace*{2.5cm} > ~~ \sum_i \bOmega^{(i)}(\bar{\bW},\bb,\lambda \bI) +  n \left[ \sum_j \log \left(\lambda +  \| \bar{\bw}_{\cdot j} \|_2^2  \right) + (d-\kappa) \log \lambda \right],  \nonumber
\end{eqnarray}
otherwise $\bW^{**}$ would not be a minimum of (\ref{eq:ppca_cost_separable}).  Therefore, $\bar{\bW}$ must also be a minimum of (\ref{eq:ppca_cost_separable}), from which the remaining parts of (\ref{eq:ppca_equal_cost}) immediately follows.

We next confront the arrangement of disconnected minima for part 2 of the theorem.  It is not difficult to show that (\ref{eq:ppca_cost}), and by virtue of the analysis above (\ref{eq:ppca_cost_separable}), will be uniquely minimized by $\bU$ and $\bLambda$ arising from the SVD of either $\bW^{*}$ or equivalently $\bW^{**}$.  Let $\bW^{**} = \bU \bLambda \bV^{\top}$ via such a decomposition.  So any partitioning into disconnected minimizers must come at the hands of $\bV$, which only influences the $\sum_j \log \left(\lambda + \|\bw_{\cdot j} \|_2^2  \right)$ term in (\ref{eq:ppca_cost_separable}).

At stated above, if $\bW^{**}$ is a minimum, then $\bW^{**} \bP $ must also be a minimum.  Assume for the moment that $\bW^{**}$ is full column rank.  There will obviously be $r!$ unique permutations of its columns, with $r = \mbox{rank}[\bW^{**}]$.   Moreover, any transition from some permutation $\bP'$ to another $\bP''$ will necessarily involve some non-permutation-matrix rotation $\bV$.  Given our assumption of distinct eigenvalues, this will ensure that
\begin{equation}
\tfrac{1}{\lambda} \left(\bW^{**}\right)^{\top} \bW^{**}  + \bI = \tfrac{1}{\lambda}  \bV \bLambda^2 \bV^{\top}    + \bI
\end{equation}
is non-diagonal.  While this will not increase (\ref{eq:ppca_cost}), it \emph{must} increase (\ref{eq:ppca_cost_separable}) when diagonalized by Hadamard's inequality.  Therefore every permutation will reflect a distinct, disconnected minimizer.  If $\bW^{**}$ also has $\kappa - r$ zero-valued columns, then the resulting number of unique permutations increases to $\tfrac{\kappa!}{\kappa -r}$ by standard rules of combinatorics.

Finally, part 3 of the theorem follows directly from part 2:  Given that any minimizer of (\ref{eq:ppca_cost_separable}) must be of the form $\bU \bLambda \bP$, then there cannot be more than $r$ nonzero columns.  In contrast, for (\ref{eq:ppca_cost}) we may apply any arbitrary rotation to $\bW^*$, and hence all columns can be nonzero even if the rank is smaller than $\kappa$.

%========================================================
\section{Proof of Corollary \ref{cor:affine_vae_model}}
\label{sec:proof_corollary1}

Although the analysis that follows will hold for any $\bb$, for simplicity we simply fix $\bb = \bar{\bx} \triangleq \tfrac{1}{n} \sum_{i=1}^n \bx^{(i)}$, i.e., the value which minimizes the VAE cost under the current setting.  The maximum likelihood estimate of the data covariance then becomes  $\bSigma_{ML} = \tfrac{1}{n} \sum_{i=1}^n \left(\bx^{(i)} - \bar{\bx} \right) \left(\bx^{(i)} - \bar{\bx} \right)^{\top}$.  With these definitions in mind, using a standard trace identity the separable VAE cost from (\ref{eq:ppca_cost_separable}) can be equivalently expressed
\begin{equation} \label{eq:vae_cost_tot_affine4_sup}
\calL(\theta,\bphi)   \equiv    \mbox{tr}\left[ \bSigma_{ML} \left( \bW \bW^{\top}  + \lambda \bI\right)^{-1} \right] + \sum_{j=1}^{\kappa}  \log \left( \| \bw_{\cdot j}\|_2^2 + \lambda \right) + (d-\kappa) \log \lambda.
\end{equation}
Because $\log \left( \| \bw_{\cdot j}\|_2^2 + \lambda \right)$ is a concave non-decreasing function of $\| \bw_{\cdot j}\|_2^2$, it can be expressed via the variational form
\begin{eqnarray}
\log \left( \| \bw_{\cdot j}\|_2^2 + \lambda \right) & = & \min_{\omega_j \geq 0} \left\{ \omega_j \| \bw_{\cdot j}\|_2^2 - h^*(\omega_j) \right\},
\end{eqnarray}
where $h^*(\omega)$ denotes the concave conjugate function \cite{Boyd04} of $h(u) \triangleq \log(u + \lambda)$, $u \geq 0$.  This formulation produces a strict upper bound once we drop the minimization, i.e.,
\begin{eqnarray}
\log \left( \| \bw_{\cdot j}\|_2^2 + \lambda \right) & \geq &  \omega_j \| \bw_{\cdot j}\|_2^2 - h^*(\omega_j), ~\forall \omega_j \geq 0.
\end{eqnarray}
Now assume for the moment that $\bW'$ is some local minimum of (\ref{eq:vae_cost_tot_affine4_sup}) and that, for all $j$, we have that
\begin{eqnarray}
\log \left( \| \bw'_{\cdot j}\|_2^2 + \lambda \right) & = &  \omega'_j \| \bw'_{\cdot j}\|_2^2 - h^*(\omega'_j)
\end{eqnarray}
for some corresponding non-negative $\omega'_j$ values.  By the above variational construction we are guaranteed that such $\omega'_j$ will always exist.  If $\bW'$ is truly a local minimum, then it must also be a local minimum of the upper bound
\begin{eqnarray} \label{eq:vae_cost_tot_affine5_sup}
\bar{\calL}(\btheta,\bphi)
\triangleq  &&  \\
&& \hspace*{-1.2cm} \mbox{tr}\left[ \bSigma_{ML} \left( \bW \bW^{\top}  + \lambda \bI\right)^{-1} \right] + \sum_{j=1}^\kappa  \left[ \omega'_j \| \bw_{\cdot j}\|_2^2 - h^*(\omega'_j) \right] + (d-\kappa) \log \lambda. \nonumber
\end{eqnarray}
Note that at the point $\bW = \bW'$, $\bar{\calL}(\btheta,\bphi) = \calL(\btheta,\bphi)$, and yet by construction $\bar{\calL}(\btheta,\bphi) \geq \calL(\btheta,\bphi)$ everywhere.  Therefore if the bound is not minimized at this point, then $\bW'$ cannot be a minimum to $\calL(\btheta,\bphi)$.

Using the standard singular value decomposition, let $\bW = \bU \bLambda \bV^{\top}$, where $\bU$ and $\bV$ are square orthonormal matrices and $\bLambda$ is a positive semi-definite diagonal matrix.  Any minimum of $\bar{\calL}(\btheta,\bphi)$ or $\calL(\btheta,\bphi)$ with respect to $\bW$ must also be a minimum with respect to $\bU$, $\bLambda$, and $\bV$, otherwise we could smoothly alter one of these components to smoothly change $\bW$ and reduce the cost.  Now denote $\bOmega$ as a zero-valued matrix with each $\omega'_j$ placed in the $j$-th diagonal position.  Then we have that
\begin{equation}
\sum_{j=1}^\kappa  \omega'_j \| \bw_{\cdot j}\|_2^2  = \mbox{tr}\left[\bW^{\top} \bW \bOmega  \right] = \mbox{tr}\left[\bV \bLambda^2 \bV^{\top} \bOmega  \right].
\end{equation}
Because $\bW \bW^{\top} = \bU \bLambda^2 \bU^{\top}$ is independent of $\bV$, minimization of $\bar{\calL}(\btheta,\bphi)$ need only involve the term $\mbox{tr}\left[\bV \bLambda^2 \bV^{\top} \bOmega  \right]$.  Therefore, a necessary condition for any true local minimum is that it must occur at a stationary point of the reduced problem
\begin{equation} \label{eq:contrained_quadratic_problem}
\min_{\bV} ~ \mbox{tr}\left[\bV \bLambda^2 \bV^{\top} \bOmega^2  \right], ~~~\mbox{s.t. } \bV^{\top} \bV = \bI.
\end{equation}
Using results from \cite{Brockett1989}, it can be shown that the stationary points of (\ref{eq:contrained_quadratic_problem}) must occur when $\bV$ is a permutation matrix, at least assuming diagonal elements of $\bLambda^2$ and $\bOmega$ are distinct; however, a simple continuity argument can be used to extend to the general case.

Proceeding further, if $\bV$ must be a permutation matrix, then at any local minimum with respect to $\bV$, it must be that $\| \bw_{\cdot j}\|_2^2 = \lambda_j^2$, where $\lambda_j$ is the $j$-th diagonal element of $\bLambda$.  From this observation we may infer that
\begin{eqnarray}
\min_{\omega_j \geq 0} \left\{ \omega_j \| \bw_{\cdot j}\|_2^2 - h^*(\omega_j) \right\} & = & \min_{\omega_j \geq 0} \left\{ \omega_j \lambda_j^2 - h^*(\omega_j) \right\} \nonumber \\
& = & \log \left| \bLambda^2 + \lambda \bI \right| ~~ \equiv ~~ \log \left| \bU \bLambda^2 \bU^{\top} + \lambda \bI \right|.
\end{eqnarray}
Since $\bW \bW^{\top} = \bU \bLambda^2 \bU^{\top}$, it then follows that any local minimum with respect to $\bU$ and $\bLambda^2$ must also be a local minimum of the revised cost
\begin{equation}
\widetilde{\calL}(\btheta,\bphi)  \triangleq    \mbox{tr}\left[ \bSigma_{ML} \left( \bU \bLambda^2 \bU^{\top}  + \lambda \bI\right)^{-1} \right] + \log \left| \bU \bLambda^2 \bU^{\top} + \lambda \bI \right|.
\end{equation}
It has been demonstrated in \cite{Tipping1999} that all local minima of this simplified objective are global, with $\bU \bLambda$ spanning the principal subspace of $\bSigma_{ML}$ associated with eigenvalues larger than $\lambda$.  Additionally, by Hadamard's inequality \cite{Garling07} we have
\begin{eqnarray}
\log \left| \bU \bLambda^2 \bU^{\top} + \lambda \bI \right| & = & \log \left| \tfrac{1}{\lambda} \bW^{\top} \bW  +  \bI \right| + d\log \lambda  \\
& \leq & \sum_{j=}^{\kappa} \left(1+ \tfrac{1}{\lambda} \| \bw_{\cdot j}\|_2^2 \right) + d \log \lambda  =  \sum_j \log \left( \| \bw_{\cdot j}\|_2^2 + \lambda \right) + (d-\kappa) \log \lambda.  \nonumber
\end{eqnarray}
Therefore it must also be true that $\widetilde{\calL}(\btheta,\bphi) \leq \calL(\btheta,\bphi)$, and so these global minima must also be global minima to our original objective.

\section{Proof of Theorem~\ref{thm:vae_and_rpca}}
\label{sec:proof_theorem2}

For convenience we will adopt the notation $f(\alpha) = \mathrm{O}(h(\alpha))$ to indicate that there exists a positive $\bar{\alpha}$ and some constant $C$ independent of $\alpha$ such that $|f(\alpha)| < C h(\alpha)$ for all $\alpha \in (0,\bar{\alpha}]$.  Similarly, we use $f(\alpha) = \Omega(h(\alpha))$ to convey that $|f(\alpha)| > C h(\alpha)$ under equivalent conditions.  We then say $f(\alpha) = \Theta(h(\alpha))$ iff $f(\alpha) = \mathrm{O}(h(\alpha))$ and $f(x) = \Omega(h(\alpha))$.  Additionally, if the input argument to one of these expressions is a vector, the result is understood to apply element-wise.

The basic high-level strategy here is as follows:  We first present a candidate solution that satisfies (\ref{eq:optimal_conditions}) and carefully quantify the achievable objective function value for $\alpha \in (0,\bar{\alpha}]$, and $\bar{\alpha}$ small.  We then analyze a lower bound on the VAE cost and demonstrate that no solution can do significantly better, namely, any solution that can match the performance of our original proposal must necessarily also satisfy (\ref{eq:optimal_conditions}).  Given that this is a lower bound, this implies that no other solution can both minimize the VAE objective and not satisfy (\ref{eq:optimal_conditions}).  We now proceed to the details.

Define $\bmu_z^{(i)} \triangleq \bmu_z\left(\bx^{(i)}; \bphi \right)$ and $\bSigma_z^{(i)} \triangleq \bSigma_z\left(\bx^{(i)} ; \bphi \right)$. We first note that if $\bz = \bmu_z^{(i)} + \bS_z^{(i)} \bepsilon$, with $\bS_z^{(i)}$ satisfying $\bSigma_z^{(i)} = \bS_z^{(i)} \left(\bS_z^{(i)}\right)^{\top}$, and $\bepsilon \sim p(\bepsilon) = \calN(\bepsilon ; {\bf 0}, \bI)$, then $\bz \sim q_{\tiny \bphi}\left(\bz|\bx^{(i)} \right)$.  With this reparameterization and
\begin{eqnarray}
\bmu_x^{(i)} & \triangleq & \bW \bmu_z^{(i)} + \bW \bS_z^{(i)} \bepsilon, \nonumber \\
\mbox{diag}[\bSigma_x^{(i)}] & \triangleq & \nu \left( \bmu_z^{(i)} + \bS_z^{(i)} \bepsilon ; \btheta \right) \mbox{ for some function } \nu \nonumber \\
\bmu_z^{(i)} & \triangleq & f(\bx^{(i)} ; \bphi) \mbox{ for some function } f    \\
\bS_z^{(i)} & \triangleq &  g(\bx^{(i)}; \bphi) \mbox{ for some function } g,  \nonumber
\end{eqnarray}
the equivalent VAE objective becomes
\begin{eqnarray} \label{eq:vae_cost_part_affine_sup}
\calL(\btheta,\bphi) &  =  & \sum_i \left\{ \mathbb{E}_{p\left(\bepsilon \right)} \left[ \left( \bx^{(i)} - \bW \bmu_z^{(i)} - \bW \bS_z^{(i)} \bepsilon  \right)^{\top} \left( \bSigma_x^{(i)}  \right)^{-1} \left( \bx^{(i)} - \bW \bmu_z^{(i)} - \bW \bS_z^{(i)} \bepsilon  \right)  \right]   \right. \nonumber \\
& & +  ~~ \left.  \mathbb{E}_{p\left(\bepsilon \right)} \left[ \log \left| \bSigma_x^{(i)} \right| \right] + \mbox{tr}\left[ \bSigma_z^{(i)} \right] - \log \left| \bSigma_z^{(i)}   \right| + \| \bmu_z^{(i)} \|_2^2  \right\}
\end{eqnarray}
when $\bb = {\bf 0}$ as stipulated.\footnote{The extension to arbitrary $\bb$ is trivial but clutters the presentation.}  For now assume that $\kappa$, the dimension of the latent $\bz$, satisfies $\kappa = \mbox{rank}[\bU]$ (later we will relax this assumption).

\subsection{A Candidate Solution} \label{sec:candidate_solution}
Here we consider a candidate solution that, by design, satisfies (\ref{eq:optimal_conditions}).  For the encoder parameters we choose
\begin{equation} \label{eq:z_moments_sup}
\hat{\bmu}_z^{(i)}  =  \bpi^{(i)}, ~~~~ \hat{\bSigma}_z^{(i)}  = \alpha \bI.
\end{equation}
where $\alpha$ is a non-negative scalar and $\bpi^{(i)}$ is defined in conjunction with a matrix $\bPsi$ such that
\begin{eqnarray}
\mbox{supp}_{\alpha} \left[\bx^{(i)} - \bPsi \bpi^{(i)}  \right] & = & \mbox{supp}\left[\bs^{(i)} \right] \nonumber \\
\mbox{span}\left[\bU \right] & = & \mbox{span}\left[ \bPsi  \right].
\end{eqnarray}
All quantities in (\ref{eq:z_moments_sup}) can be readily computed via $\bX$ applied to an encoder module provided that $\kappa = \mbox{dim}[\bz] = \mbox{rank}\left[\bU\right]$ as stipulated, and sufficient representational complexity for $\bmu_z$ and $\bSigma_z$.  Additionally, for the encoder we only need to define the posterior moments at specific points $\bx^{(i)}$, hence the indexing via $i$ in (\ref{eq:z_moments_sup}).

In contrast, for the decoder we consider the solution defined over any $\bz$ given by
\begin{eqnarray} \label{eq:x_moments_sup}
\hat{\bW} & = & \bPsi \nonumber \\
\hat{\bmu}_x & = & \hat{\bW} \bz \nonumber \\
\mbox{diag}\left[\hat{\bSigma}_x \right] & = & \bLambda^{(h_{\pi}(\bz))},
\end{eqnarray}
where  $\bLambda^{(i)} \in \mathbb{R}^{d\times d}$ is a diagonal matrix with
\begin{equation}
\left[ \bLambda^{(i)} \right]_{jj}  =   \left\{ \begin{array}{ll}
\alpha, & \mbox{if}~~s_j^{(i)} = 0,  \\
 1, & \mbox{otherwise}, ~~~ \forall j.
\end{array} \right.
\end{equation}
and $h_{\pi} : \mathbb{R}^{\kappa} \rightarrow \{1,\ldots,n\}$ is a function satisfying
\begin{equation}
h_{\pi}(\bz) \triangleq \arg \min_{i \in \{1,\ldots,n\} } \| \bz - \bpi^{(i)} \|_2.
\end{equation}
Again, given sufficient capacity, this function can always be learned by the decoder such that (\ref{eq:x_moments_sup}) is computable for any $\bz$.  Given these definitions, then the index-specific moments $\hat{\bmu}_x^{(i)}$ and $\hat{\bSigma}_x^{(i)}$ are of course reduced to functions of $\bepsilon$ given by
\begin{eqnarray}
\hat{\bmu}_x^{(i)} & = & \hat{\bmu}_x \left(\hat{\bmu}_z^{(i)} + \hat{\bS}_z^{(i)} \bepsilon; \btheta \right) \nonumber \\
\hat{\bSigma}_x^{(i)} & = & \hat{\bSigma}_x \left(\hat{\bmu}_z^{(i)} + \hat{\bS}_z^{(i)} \bepsilon ; \btheta \right) .
\end{eqnarray}
.

% rounds its vector-valued argument to the nearest integer in element-wise fashion.

We next analyze the behavior of (\ref{eq:vae_cost_part_affine_sup}) at this specially parameterized solution as $\bar{\alpha}$ becomes small, in which case by design all covariances will be feasible by design.  For this purpose, we first consider the integration across all cases where $\hat{\bSigma}_x^{(i)}$ does not reflect the correct support, meaning $\bepsilon \notin \calS^{(i)}$,
where
\begin{equation}
\calS^{(i)} \triangleq \left\{ \bepsilon : ~~  \left[ \bSigma_x^{(i)} \right]_{jj} = \alpha ~~\mbox{iff} ~~ s_j^{(i)} = 0, ~~ \forall j \right\}.
\end{equation}
With this segmentation in mind, the VAE objection naturally partitions as
\begin{equation}
\calL(\btheta,\bphi) = \sum_i \left\{ \calL^{(i)}(\btheta,\bphi; \epsilon \notin \calS^{(i)}) +  \calL^{(i)}(\btheta,\bphi; \epsilon \in \calS^{(i)}) \right\},
\end{equation}
where $\calL^{(i)}(\btheta,\bphi; \epsilon \notin \calS^{(i)})$ denotes the cost for the $i$-th sample when integrated across those samples not in $\calS^{(i)}$, and $\calL^{(i)}(\btheta,\bphi; \epsilon \in \calS^{(i)})$ is the associated complement.

\subsection{Evaluation of $\calL^{(i)}(\btheta,\bphi; \epsilon \notin \calS^{(i)})$}
First we define
\begin{equation}
\rho =  \min_{i,j \in \{1,\ldots,n\}, i \neq j } \tfrac{1}{2}\| \bpi^{(i)} - \bpi^{(j)} \|_2,
\end{equation}
which is just half the minimum distance between any two distinct coefficient expansions.  If any $\bz$ is within this distance of  $\bpi^{(i)}$, it will necessarily be quantized to this value per our previous definitions.  Therefore if $\|\hat{\bS}_z^{(i)} \bepsilon\|_2 < \rho$, we are guaranteed that the correct generating support pattern will be mapped to $\hat{\bSigma}_x^{(i)}$, and so it follows that
\begin{equation}
P\left(\bepsilon \notin \calS^{(i)} \right)) \leq P\left( \left\| \hat{\bS}_z^{(i)} \bepsilon\right\|_2 > \rho \right) = P( \|\sqrt{ \alpha } \bepsilon \|_2 > \rho )
\end{equation}
at our candidate solution.  We also make use of the quantity
\begin{equation}
\eta \triangleq \max_{i \in \{1,\ldots,n\} } \| \bx^{(i)} - \bPsi \bpi^{(i)}\|_2^2,
\end{equation}
which represents the maximum data-fitting error.  Then for the $i$-th sample we have
\begin{eqnarray} \label{eq:vae_cost_part_affine_sup2}
& & \hspace*{-1.5cm} \calL^{(i)}(\btheta,\bphi; \epsilon \notin \calS^{(i)}) \nonumber \\
& = &  \int_{\bepsilon \notin \calS^{(i)}  }  \left[ \left( \bx^{(i)} - \hat{\bW} \hat{\bmu}_z^{(i)} - \hat{\bW} \hat{\bS}_z^{(i)} \bepsilon  \right)^{\top} \left( \hat{\bSigma}_x^{(i)}  \right)^{-1} \left( \bx^{(i)} - \hat{\bW} \hat{\bmu}_z^{(i)} - \hat{\bW} \hat{\bS}_z^{(i)} \bepsilon  \right)  \right.    \nonumber \\
& & +  ~~ \left.   \log \left| \hat{\bSigma}_x^{(i)} \right|  + \mbox{tr}\left[ \hat{\bSigma}_z^{(i)} \right] - \log \left| \hat{\bSigma}_z^{(i)}   \right| + \| \hat{\bmu}_z^{(i)} \|_2^2  \right] p(\bepsilon) d \bepsilon \nonumber \\
&  \leq  & \int_{\|\sqrt{ \alpha } \bepsilon \|_2 > \rho }  \left[ \left( \bx^{(i)} - \hat{\bW} \hat{\bmu}_z^{(i)} - \hat{\bW} \hat{\bS}_z^{(i)} \bepsilon  \right)^{\top} \left( \hat{\bSigma}_x^{(i)}  \right)^{-1} \left( \bx^{(i)} - \hat{\bW} \hat{\bmu}_z^{(i)} - \hat{\bW} \hat{\bS}_z^{(i)} \bepsilon  \right)  \right.    \nonumber \\
& & +  ~~ \left.   \log \left| \hat{\bSigma}_x^{(i)} \right|  + \mbox{tr}\left[ \hat{\bSigma}_z^{(i)} \right] - \log \left| \hat{\bSigma}_z^{(i)}   \right| + \| \hat{\bmu}_z^{(i)} \|_2^2  \right] p(\bepsilon) d \bepsilon \nonumber \\
& \leq & \int_{\|\sqrt{ \alpha } \bepsilon \|_2 > \rho }   \left[ \tfrac{1}{\alpha} \left( \bx^{(i)} - \bPsi \bpi^{(i)} - \sqrt{\alpha} \bPsi \bepsilon  \right)^{\top}  \left( \bx^{(i)} - \bPsi \bpi^{(i)} - \sqrt{\alpha} \bPsi \bepsilon  \right)  \right.    \nonumber \\
& & +  ~~ \left.  \kappa \alpha - \kappa \log \alpha + \| \bpi^{(i)} \|_2^2  \right]  p(\bepsilon) d \bepsilon,
\end{eqnarray}
where the second inequality comes from setting $\hat{\bSigma}_x^{(i)} = \alpha \bI$ (its smallest possible value) in the inverse term and $\hat{\bSigma}_x^{(i)} = \bI$ (its largest value) in the log-det term.  Next, given that
\begin{eqnarray}
\| \bx^{(i)} - \bPsi \bpi^{(i)}\|_2^2 & \leq & \eta \nonumber \\
\int_{\|\sqrt{ \alpha } \bepsilon \|_2 > \rho } \left( \bpi^{(i)} \right)^{\top} \bPsi^{\top} \bPsi \bepsilon \cdot p(\bepsilon) d \bepsilon & = & 0  \\
\int_{\|\sqrt{ \alpha } \bepsilon \|_2 > \rho } \| \bPsi \bepsilon \|_2^2  p(\bepsilon) d \bepsilon & \leq & \mbox{tr}\left[\bPsi^{\top} \bPsi \right], \nonumber
\end{eqnarray}
it follows that the bound from (\ref{eq:vae_cost_part_affine_sup2}) can be further reduced via
\begin{eqnarray}
\calL^{(i)}(\btheta,\bphi; \epsilon \notin \calS^{(i)}) & \leq & \mbox{tr}\left[\bPsi^{\top} \bPsi \right] + \int_{\|\sqrt{ \alpha } \bepsilon \|_2 > \rho }  \left[ \tfrac{1}{\alpha} \eta  + \kappa \alpha - \kappa \log \alpha + \| \bpi^{(i)} \|_2^2  \right] p(\bepsilon) d \bepsilon \nonumber \\
& = &  \Theta(1) + \left[ \tfrac{1}{\alpha} \eta  + \kappa \alpha - \kappa \log \alpha + \| \bpi^{(i)} \|_2^2  \right] \int_{\|\sqrt{ \alpha } \bepsilon \|_2 > \rho }  p(\bepsilon) d \bepsilon \nonumber \\
& \leq &  \Theta(1) + \left[ \tfrac{1}{\alpha} \eta  + \kappa \alpha - \kappa \log \alpha + \| \bpi^{(i)} \|_2^2  \right] \tfrac{\alpha }{\rho^2} \nonumber \\
& = & \Theta(1) + \Theta(\alpha^2) - \Theta(\alpha \log \alpha) \nonumber \\
& = & \Theta(1) ~~~ \mbox{as} ~  \alpha \rightarrow 0,
\end{eqnarray}
where the second inequality holds based on the vector version of Chebyshev's inequality, which ensures that
\begin{equation}
\int_{\|\sqrt{ \alpha } \bepsilon \|_2 > \rho }  p(\bepsilon) d \bepsilon = P(\|\sqrt{ \alpha } \bepsilon \|_2 > \rho) \leq \tfrac{\alpha}{\rho^2}.
\end{equation}
Clearly then, as $\alpha$ becomes small, we have established that
\begin{equation} \label{eq:cost_val_part_1_sup}
\calL^{(i)}(\btheta,\bphi; \epsilon \notin \calS^{(i)}) \rightarrow O\left( 1 \right).
\end{equation}

\subsection{Evaluation of $\calL^{(i)}(\btheta,\bphi; \epsilon \in \calS^{(i)})$}

In analyzing $\calL^{(i)}(\btheta,\bphi; \epsilon \in \calS^{(i)})$, we note that
\begin{eqnarray}
&& \hspace*{-2.0cm} \int_{\bepsilon \in \calS^{(i)}  }   \left( \bx^{(i)} - \hat{\bW} \hat{\bmu}_z^{(i)} - \hat{\bW} \hat{\bS}_z^{(i)} \bepsilon  \right)^{\top} \left( \hat{\bSigma}_x^{(i)}  \right)^{-1} \left( \bx^{(i)} - \hat{\bW} \hat{\bmu}_z^{(i)} - \hat{\bW} \hat{\bS}_z^{(i)} \bepsilon  \right)  p(\bepsilon) d \bepsilon \nonumber \\
& \leq & \int  \left( \bx^{(i)} - \bPsi \bpi^{(i)} -\sqrt{\alpha} \bPsi \bepsilon \right)^{\top} \left( \bLambda^{(i)}  \right)^{-1} \left( \bx^{(i)} - \bPsi \bpi^{(i)} - \sqrt{\alpha} \bPsi \bepsilon \right)  p(\bepsilon) d \bepsilon \nonumber \\
& \leq & \int  \left( \bx^{(i)} - \bPsi \bpi^{(i)}  \right)^{\top} \left( \bLambda^{(i)}  \right)^{-1} \left( \bx^{(i)} - \bPsi \bpi^{(i)}  \right)  p(\bepsilon) d \bepsilon + \mbox{tr}\left[  \bPsi^{\top} \bPsi \right] \nonumber \\
& \leq & \eta + \mbox{tr}\left[  \bPsi^{\top} \bPsi \right] \nonumber \\
& = & \Theta(1)
\end{eqnarray}
given the alignment of $\bLambda^{(i)}$ with zero-valued elements in $\bx^{(i)} - \bPsi \bpi^{(i)}$.  Furthermore, the remaining terms in $\calL^{(i)}(\btheta,\bphi; \epsilon \in \calS^{(i)})$ are independent of $\bepsilon$ giving
\begin{eqnarray}
&& \hspace*{-1.0cm} \int_{\bepsilon \in \calS^{(i)}  } \left[ \log \left| \hat{\bSigma}_x^{(i)} \right| + \mbox{tr} \left[ \hat{\bSigma}_z^{(i)}   \right] - \log \left| \hat{\bSigma}_z^{(i)}   \right|   + \| \hat{\bmu}_z^{(i)} \|_2^2\right] p(\bepsilon) d \bepsilon \nonumber \\
& = & \left[ \log \left| \bLambda^{(i)} \right| + \kappa \alpha - \kappa \log \alpha   + \| \bpi^{(i)} \|_2^2\right]  \int_{\|\sqrt{ \alpha } \bepsilon \|_2 < \rho  } p(\bepsilon) d \bepsilon \nonumber \\
& = & \left[ (r^{(i)} - \kappa) \log \alpha + \kappa \alpha    + \| \bpi^{(i)} \|_2^2\right]  \int_{\|\sqrt{ \alpha } \bepsilon \|_2 < \rho  } p(\bepsilon) d \bepsilon \nonumber \\
 & = & \left[ (r^{(i)} - \kappa) \log \alpha \right] \int_{\|\sqrt{ \alpha } \bepsilon \|_2 < \rho  } p(\bepsilon) d \bepsilon + O(\alpha) + O(1),
\end{eqnarray}
where
\begin{equation}
r^{(i)} \triangleq \left| \left\{ j : \Lambda_{jj}^{(i)} = \alpha \right\} \right| = d - \| \bs^{(i)} \|_0.
\end{equation}
Therefore, since $\int_{\|\sqrt{ \alpha } \bepsilon \|_2 < \rho  } p(\bepsilon) d \bepsilon \rightarrow 1$ as $\alpha$ becomes small, we may conclude that
\begin{equation}  \label{eq:cost_val_part_2_sup}
\calL^{(i)}(\btheta,\bphi; \epsilon \in \calS^{(i)}) \rightarrow \left( d - \kappa - \| \bs^{(i)} \|_0 \right) \log \alpha + O(1).
\end{equation}

\subsection{Compilation of Candidate Solution Cost}

After combining (\ref{eq:cost_val_part_1_sup}) and (\ref{eq:cost_val_part_2_sup}) across all $i$ we find that
\begin{equation} \label{eq:upper_bound_on_cost_sup}
\calL(\btheta,\bphi ) \rightarrow \sum_i \left( d - \kappa - \| \bs^{(i)} \|_0 \right) \log \alpha + O(1)
\end{equation}
for any $\alpha \in (0,\bar{\alpha}]$ as $\bar{\alpha}$ becomes small.  If $d > \kappa + \| \bs^{(i)} \|_0$, then this expression will tend towards minus infinity, indicative of an objective value that is unbounded from below, certainly a fertile region for candidate minimizers.  Note that per the theorem statement, $\bL = \bU \bV$ and $\bS$ must represent a unique feasible solution to
\begin{equation}
\min_{\bL,\bS} ~~ d \cdot \mbox{rank}[\bL] + \| \bS \|_0 ~~~ \mbox{s.t.} ~~ \bX = \bL + \bS.
\end{equation}
Given that each column $\bx^{(i)}$ has $d$ degrees of freedom, then with $\bU$ fixed there will be an infinite number of feasible solutions  $ \bx^{(i)} = \bU \bv^{(i)} + \bs^{(i)}$ such that $\mbox{dim}[\bv^{(i)}] + \| \bs^{(i)}\|_0 = \kappa + \| \bs^{(i)}\|_0 > d$ and a combinatorial number such that $k + \| \bs^{(i)}\|_0 = d$.  Therefore for uniqueness we require that $k + \| \bs^{(i)} \|_0 < d$, so it follows that indeed $\calL(\btheta,\bphi )$ will be unbounded from below as $\bar{\alpha}$ and therefore $\alpha$ becomes small, with cost given by (\ref{eq:upper_bound_on_cost_sup}) as a candidate solution satisfying the conditions of the theorem.

Of course it still remains possible that some other candidate solution could exist that violates one of these conditions and yet still achieves (\ref{eq:upper_bound_on_cost_sup}) or an even lower cost.  We tackle this issue next.  For this purpose our basic strategy will be to examine a \emph{lower} bound on $\calL(\btheta,\bphi )$ and show that essentially any candidate solution violating the theorem conditions will be worse than (\ref{eq:upper_bound_on_cost_sup}).

\subsection{Evaluation of Other Candidate Solutions} \label{sec:other_candidate_solutions_sup}

To begin, we first observe that if granted the flexibility to optimize $\bSigma_x^{(i)}$ independently over all values of $\bepsilon$ inside the integral for computing $\calL(\btheta,\bphi )$, we immediately obtain a rigorous lower bound.\footnote{Note that this is never exactly achievable in practice, even with an infinite capacity network for computing $\bSigma_x^{(i)}$, since it would require a unique network for each data sample; however, it nonetheless serves as a useful analysis tool.} For this purpose we must effectively solve decoupled problems of the form
\begin{equation} \label{eq:simple_opt_prob_sup}
\inf_{\gamma > \alpha} \tfrac{c}{\gamma} + \log \gamma,
\end{equation}
to which the optimal solution is just
\begin{equation}
\gamma^* = \xi_{\alpha} (x) \triangleq [c - \alpha]_+ + \alpha,
\end{equation}
 where the operator $[\cdot]_+$ retains only the positive part of its argument, setting negative values to zero. Plugging this solution back into (\ref{eq:simple_opt_prob_sup}), we find that
\begin{equation} \label{eq:simple_opt_prob2_sup}
\inf_{\gamma > \alpha} \tfrac{c}{\gamma} + \log \gamma ~~ = ~~ \log \xi_{\alpha} (c) + O(1).
\end{equation}
In the context of our bound, this leads to

\begin{eqnarray} \label{eq:vae_cost_part_affine_lower_bound_sup}
\calL(\btheta,\bphi) &  \geq  & \sum_i \left\{ \mathbb{E}_{p\left(\bepsilon \right)} \left[ \sum_j \log \xi_{\alpha} \left( \left[ x_j^{(i)} - \bw_{j\cdot} \bmu_z^{(i)} - \bw_{j\cdot} \bS_z^{(i)} \bepsilon \right]^2 \right)  \right]   \right. \nonumber \\
& & \left.  + ~~\mbox{tr}\left[ \bSigma_z^{(i)} \right] - \log \left| \bSigma_z^{(i)}   \right| + \| \bmu_z^{(i)} \|_2^2  \right\} + O(1).
\end{eqnarray}
From this expression, it is clear that the lowest objective value we could ever hope to obtain cannot involve arbitrarily large values of $\bSigma_z^{(i)}$ and $\bmu_z^{(i)}$ since the respective trace and quadratic terms grow faster than log-det terms.  Likewise $\bmu_z^{(i)}$ cannot be unbounded for analogous reasons.  Therefore, optimal solutions to  (\ref{eq:vae_cost_part_affine_lower_bound_sup}) that will be unbounded from below must involve the first term becoming small, at least over a range of $\bepsilon$ values with significant probability measure.  Although the required integral admits no closed-form solution, we can simplify things further using refinements of the above bound.

For this purpose consider any possible candidate solution $\hat{\bW} = \bPsi$ and $\hat{\bmu}_z^{(i)} = \bpi^{(i)}$ (not necessarily one that coincides with $\bU$ and the optimal subspace), and define
\begin{equation}
\Delta_{\alpha}^{(i)}(\bPsi,\bpi ) \triangleq \mbox{supp}_{\alpha} \left[\bx^{(i)} - \bPsi \bpi  \right].
\end{equation}
Without loss of generality we also specify that
\begin{equation}
\bS_z^{(i)} \triangleq \bXi^{(i)} \bD^{(i)},
\end{equation}
where $\bXi^{(i)} \in \mathbb{R}^{d\times \kappa}$ has orthonormal columns and $\bD^{(i)}$ is a diagonal matrix with
\begin{equation}
\left[ \bD^{(i)} \right]_{kk} = \xi_{\sqrt{\alpha}} \left( \sigma_k^{(i)} \right),
\end{equation}
and $\bsigma^{(i)} = [\sigma_1^{(i)}, \ldots, \sigma_{\kappa}^{(i)}]^{\top} \in \mathbb{R}^{\kappa}_+$ is an arbitrary non-negative vector.  Any general $\bSigma_z^{(i)} = \bS_z^{(i)} \left( \bS_z^{(i)} \right)^{\top}$, with singular values bounded by $\alpha$,  is expressible via this format.  We then reexpress (\ref{eq:vae_cost_part_affine_lower_bound_sup}) as
\begin{eqnarray} \label{eq:vae_cost_part_affine_lower_bound2_sup}
\calL(\btheta,\bphi) &  \geq  & \sum_i \left\{ \mathbb{E}_{p\left(\bepsilon \right)} \left[ \sum_{j \in \Delta_{\alpha}^{(i)}\left(\bPsi,\bpi^{(i)} \right)} \log \xi_{\alpha} \left( \left[ x_j^{(i)} - \bpsi_{j\cdot} \bpi^{(i)} - \bpsi_{j\cdot} \bXi^{(i)} \bD^{(i)} \bepsilon \right]^2 \right)  \right]   \right. \nonumber \\
 &  +  & \mathbb{E}_{p\left(\bepsilon \right)} \left[ \sum_{j \notin \Delta_{\alpha}^{(i)}\left(\bPsi,\bpi^{(i)} \right)} \log \xi_{\alpha} \left( \left[ O(\alpha) + \bpsi_{j\cdot} \bXi^{(i)} \bD^{(i)} \bepsilon \right]^2 \right)  \right]  \nonumber   \\
& + & \left.  \mbox{tr}\left[ \bXi^{(i)} \left( \bD^{(i)} \right)^2 \left( \bXi^{(i)} \right)^{\top} \right] - \log \left| \bXi^{(i)} \left( \bD^{(i)} \right)^2 \left( \bXi^{(i)} \right)^{\top}   \right| + \| \bpi^{(i)} \|_2^2  \right\} + O(1), \nonumber \\
& = & \sum_i \left\{ \mathbb{E}_{p\left(\bepsilon \right)} \left[ \sum_{j \in \Delta_{\alpha}^{(i)}\left(\bPsi,\bpi^{(i)} \right)} \log \xi_{\alpha} \left( \left[ x_j^{(i)} - \bpsi_{j\cdot} \bpi^{(i)} - \sum_k \bar{\psi}_{jk} \cdot \xi_{\sqrt{\alpha}} \left( \sigma_k^{(i)}  \right) \cdot \epsilon_k \right]^2 \right)  \right]   \right. \nonumber \\
 &  +  & \mathbb{E}_{p\left(\bepsilon \right)} \left[ \sum_{j \notin \Delta_{\alpha}^{(i)}\left(\bPsi,\bpi^{(i)} \right)} \log \xi_{\alpha} \left( \left[ O(\alpha) + \sum_k \bar{\psi}_{jk}^{(i)} \cdot \xi_{\sqrt{\alpha}} \left( \sigma_k^{(i)}  \right) \cdot \epsilon_k \right]^2 \right)  \right]   \nonumber  \\
& + & \left.  \sum_k \xi_{\alpha} \left[ \left( \sigma_k^{(i)} \right)^2 \right] - \sum_k \log \xi_{\alpha} \left[ \left( \sigma_k^{(i)} \right)^2 \right] + \| \bpi^{(i)} \|_2^2  \right\} + O(1), 
\end{eqnarray}
where $\bar{\psi}_{jk}^{(i)}$ is the $k$-th element of the vector $\bpsi_{j\cdot} \bXi^{(i)}$. We can now analyze any given point $\{\bPsi, \bpi^{(i)}, \bXi^{(i)}, \bsigma^{(i)} \}_{i=1}^n$ as $\alpha$ becomes small.  The first term can be shown to be $\Theta(1)$ with all other variables fixed,\footnote{Note that $x_j^{(i)} - \bpsi_{j\cdot} \bpi^{(i)} = \Theta(1)$ for all $j \in \Delta_{\alpha}^{(i)}\left(\bPsi,\bpi^{(i)} \right)$ and $\int \log \xi_{\alpha}\left[ \left( \Theta(1) + x \right)^2 \right] \calN(x;0,\gamma) dx = \Theta(1)$ for any variance $\gamma > 0$.} leading to the revised bound
\begin{eqnarray} \label{eq:vae_cost_part_affine_lower_bound3_sup}
\calL(\btheta,\bphi) &  \geq  &  \sum_i \left\{ \mathbb{E}_{p\left(\bepsilon \right)} \left[ \sum_{j \notin \Delta_{\alpha}^{(i)}\left(\bPsi,\bpi^{(i)} \right)} \log \xi_{\alpha} \left( \left[ O(\alpha) +  \sum_k \bar{\psi}_{jk}^{(i)} \cdot \xi_{\sqrt{\alpha}} \left( \sigma_k^{(i)}  \right) \cdot \epsilon_k \right]^2 \right)  \right]  \right.   \\
& - & \left.  \sum_k \log \xi_{\alpha} \left[ \left( \sigma_k^{(i)} \right)^2 \right]   \right\} + \Theta(1), \nonumber
\end{eqnarray}
where the terms $\sum_k \xi_{\alpha} \left[ \left( \sigma_k^{(i)} \right)^2 \right]$ and $\| \bpi^{(i)} \|_2^2$ have also been absorbed into $\Theta(1)$.

Given that  % Divide by the max value to see this
\begin{equation}
\mathbb{E}_{p\left(\bepsilon \right)} \left[ \log \xi_{\alpha} \left( \left[ O(\alpha) + \ba^{\top} \bepsilon  \right]^2 \right) \right] = \log \xi_{\alpha}  \left[ \ba^{\top} \ba \right] + O\left( 1 \right) \geq \log \alpha + O\left( 1 \right)
\end{equation}
for any vector $\ba$, we have the new bound
\begin{eqnarray}
&& \hspace*{-1.0cm} \calL(\btheta,\bphi) \\
& \geq &  \sum_i \left\{ \sum_{j \notin \Delta_{\alpha}^{(i)}\left(\bPsi,\bpi^{(i)} \right)} \log \xi_{\alpha} \left( \sum_k \left[ \bar{\psi}_{jk}^{(i)} \cdot \xi_{\sqrt{\alpha}} \left( \sigma_k^{(i)}  \right) \right]^2 \right) -\sum_k \log \xi_{\alpha} \left[ \left( \sigma_k^{(i)} \right)^2 \right]   \right\} + \Theta(1). \nonumber
\end{eqnarray}
% If then we choose $\sigma_k^{(i)} = O(\alpha)$
If then we choose $\sigma_k^{(i)} = 0$ for all $i = 1,\ldots n$ and $k = 1,\ldots, \kappa$,
then
\begin{equation}
\log \xi_{\alpha} \left( \sum_k \left[ \bar{\psi}_{jk}^{(i)} \cdot \xi_{\sqrt{\alpha}} \left( \sigma_k^{(i)}  \right) \right]^2 \right) = \log \alpha + \Theta(1)
\end{equation}
and we obtain the lower bound
\begin{eqnarray} \label{eq:another_bound1_sup}
\calL(\btheta,\bphi ) & \geq & \sum_i \left( d - \kappa - \left| \Delta_{\alpha}^{(i)}\left(\bPsi,\bpi^{(i)} \right) \right| \right) \log \alpha + \Theta(1).
\end{eqnarray}
Additionally, if any set $\Delta_{\alpha}^{(i)}\left(\bPsi,\bpi^{(i)} \right)$ exists such that
\begin{equation}
\sum_i \left( d - \kappa - \left| \Delta_{\alpha}^{(i)}\left(\bPsi,\bpi^{(i)} \right) \right| \right) \leq  \sum_i \left( d - \kappa - \| \bs^{(i)} \|_0 \right),
\end{equation}
then $\bs^{(i)}$ cannot be part of the unique, feasible solution to (\ref{eq:canonical_rpca}), i.e., we could use the support pattern from each $ \Delta_{\alpha}^{(i)}\left(\bPsi,\bpi^{(i)} \right) $ to find a different feasible solution with equal or lower value of $n \cdot \mbox{rank}\left[ \bL \right] + \| \bS \|_0$, which would violate either the uniqueness or optimality of the original solution.  Therefore, we have established that with $\sigma_k^{(i)} = O(\alpha)$ for all $i$ and $k$, the resulting bound on $\calL(\btheta,\bphi )$ is essentially no better than (\ref{eq:upper_bound_on_cost_sup}), or the same bound we had before from our feasible trial solution.  Moreover, the resulting $\hat{\bW} = \bPsi$ that maximizes this bound, as well as the implicit
\begin{equation}
\hat{\bSigma}_x^{(i)}\left( \hat{\bmu}_z\left[ \bx^{(i)} \right] \right) = \mbox{diag}\left[ \left(\bx^{(i)} - \bPsi \bpi^{(i)} \right)^2\right],
\end{equation}
will necessarily satisfy (\ref{eq:optimal_conditions}).  We then only need consider whether other choices for $\sigma_k^{(i)}$ can do better.

Let $\widetilde{\bPsi}^{(i)}$ denote the the rows of $\bPsi^{(i)}$ associated with row indeces $j  \notin \Delta_{\alpha}^{(i)}\left(\bPsi,\bpi^{(i)} \right)$, meaning the indices at which we assume no sparse corruption term exists.  Additionally, define $\bB^{(i)} \triangleq \widetilde{\bPsi}^{(i)} \bXi^{(i)}$.  This implies that
\begin{eqnarray} \label{eq:final_stuff_theorem2_sup}
&& \hspace*{-2.2cm}     ~\sum_{j \notin \Delta_{\alpha}^{(i)}\left(\bPsi,\bpi^{(i)} \right)} \log \xi_{\alpha} \left( \max_k \left[ \bar{\psi}_{jk}^{(i)} \cdot \xi_{\sqrt{\alpha}} \left( \sigma_k^{(i)}  \right) \right]^2 \right) -\sum_k \log \xi_{\alpha} \left[ \left( \sigma_k^{(i)} \right)^2 \right] \\
& & \hspace*{1.0cm} = ~~ \sum_{j} \log \xi_{\alpha} \left( \max_k \left[ B_{jk}^{(i)} \cdot \xi_{\sqrt{\alpha}} \left( \sigma_k^{(i)}  \right) \right]^2 \right) -\sum_k \log \xi_{\alpha} \left[ \left( \sigma_k^{(i)} \right)^2 \right]. \nonumber
\end{eqnarray}
Contrary to our prior assumption $\bsigma^{(i)} = {\bf 0}$, now consider any solution with $\| \bsigma^{(i)} \|_0 = \beta > 0$.  For the time being, we also assume that $\bB^{(i)}$ is full column rank.  These conditions imply that
\begin{eqnarray} \label{eq:another_bound2_sup}
\sum_{j} \log \xi_{\alpha} \left( \max_k \left[ B_{jk}^{(i)} \cdot \xi_{\sqrt{\alpha}} \left( \sigma_k^{(i)}  \right) \right]^2 \right) & \geq & \left( d - \beta - \left| \Delta_{\alpha}^{(i)}\left(\bPsi,\bpi^{(i)} \right) \right| \right) \log \alpha + \Theta(1)
\end{eqnarray}
since at least $\beta$ elements of the summation over $j$ must now be order $\Theta(1)$.  By assumption we also have $\sum_k \log \xi_{\alpha} \left[ \left( \sigma_k^{(i)} \right)^2 \right] = (\kappa - \beta) \log \alpha + \Theta(1)$.  Combining with (\ref{eq:another_bound2_sup}), we see that such a solution is equivalent or worse than (\ref{eq:another_bound1_sup}).  So the former is the best we can do at any value of $\{\bPsi,\bpi^{(i)},\bXi^{(i)}, \bsigma^{(i)} \}_{i=1}^n$, provided that $\bB^{(i)}$ is full rank, and obtaining the optimal value of $\Delta_{\alpha}^{(i)}\left(\bPsi,\bpi^{(i)} \right)$ implies that (\ref{eq:optimal_conditions}) holds.

However, if $\bB^{(i)}$ is not full rank it would indeed entail that (\ref{eq:final_stuff_theorem2_sup}) could be reduced further, since a nonzero element of $\bsigma^{(i)}$ would not increase the first summation, while it would reduce the second. But if such a solution were to exist, it would violate the uniqueness assumption of the theorem statement.  To see this, note that $\mbox{rank}[\bB^{(i)}] = \mbox{rank}[\widetilde{\bPsi}^{(i)}]$ since $\bXi^{(i)}$ is orthogonal, so if the former is not full column rank, neither is the latter.  And if $\widetilde{\bPsi}^{(i)}$ is not full column rank, there will exist multiple solutions such that $\|\bx^{(i)} - \bPsi \bpi^{(i)} \|_0 = \| \bs^{(i)} \|_0$ or equivalently $\|\bx^{(i)} - \bU \bv^{(i)} \|_0 = \| \bs^{(i)} \|_0 $ in direct violation of the uniqueness clause.

Therefore to conclude, a lower bound on the VAE cost is in fact the same order as that obtainable by our original trial solution.  If this lower bound is not achieved, we cannot be at a minimizing solution, and any solution achieving this bound must satisfy (\ref{eq:optimal_conditions}).

\subsection{Generalization to Case where $\kappa > \mbox{rank}[\bU]$}

Finally, we briefly consider the case where $\kappa > \mbox{rank}[\bU] \triangleq \tau$, meaning that $\bW$ contains redundant columns that are unnecessary in producing an optimal solution to (\ref{eq:canonical_rpca}).  The candidate solution described in Section \ref{sec:candidate_solution} can be expanded via $\hat{\bW} = \left[\bPsi, ~~{\bf 0}_{[d\times (\kappa-\tau)]} \right]$, $\bmu^{(i)}_z = \left[(\bpi^{(i)})^{\top},~~ {\bf 0 }_{[1\times (\kappa-\tau)]} \right]^{\top}$, and $\hat{\bSigma}^{(i)}_z = \mbox{diag}\left[ \alpha {\bf 1}^{\top}_{[\tau\times 1]}, ~~  {\bf 1}^{\top}_{[(\kappa-\tau)\times 1]} \right]$ such that the same objective function value is obtained.

Now consider the general case where $\kappa \geq  \mbox{rank}[\hat{\bW}] > \tau$.  If we review the lower bound described in Section \ref{sec:other_candidate_solutions_sup}, with this general $\hat{\bW}$ replacing $\bPsi$, it can be shown that $\hat{\bSigma}^{(i)}_z$ will be forced to have additional diagonal elements lowered to $\alpha$, increasing the achievable objective by at least $-\log \alpha$ per sample.   The details are not especially enlightening and we omit them here for brevity.  Consequently, at any minimizer we must have $\mbox{rank}[\hat{\bW}] = \tau$.

\section{Proof of Corollary~\ref{cor:AE_behavior}}
\label{sec:proof_corollary1}

Under the stated conditions, the partially-affine VAE cost  simplifies to the function
\begin{eqnarray} \label{eq:vae_cost_no_Sigma_z_sup}
\calL(\bW,\bSigma_x,\bmu_z) & = & \sum_i \left\{  \left( \bx^{(i)} - \bW \bmu_z^{(i)}  \right)^{\top} \left( \bSigma_x^{(i)} \right)^{-1} \left( \bx^{(i)} - \bW \bmu_z^{(i)}  \right) +  \log \left| \bSigma_x^{(i)} \right| + \| \bmu_z^{(i)} \|_2^2 \right\} \nonumber \\
& = & \sum_i \left\{  \left( \bx^{(i)} - \bW \beta^{-1} \beta \bmu_z^{(i)}  \right)^{\top} \left( \bSigma_x^{(i)} \right)^{-1} \left( \bx^{(i)} - \bW \beta^{-1} \beta \bmu_z^{(i)}  \right) \right. \nonumber \\
& + & \left. \log \left| \bSigma_x^{(i)} \right| +  \beta^2 \| \bmu_z^{(i)} \|_2^2 \right\},
\end{eqnarray}
where $\beta > 0$ is an arbitrary scaler, $\bSigma_x^{(i)} \triangleq \bSigma_x\left(\bmu_z^{(i)} ; \btheta \right)$, and $\bmu_z^{(i)} \triangleq \bmu_z(\bx^{(i)}; \bphi)$.  Taking the limit as $\beta \rightarrow 0^+$, we can minimize (\ref{eq:vae_cost_no_Sigma_z_sup}) while ignoring the $\beta^2 \| \bmu_z^{(i)} \|_2^2$ regularization factor.  Consequently, we can without loss of generality consider minimization of
\begin{equation} \label{eq:vae_cost_no_Sigma_z2_sup}
\calL(\bW,\bSigma_x,\bmu_z) \equiv \sum_i \left\{  \left( \bx^{(i)} - \bW \bmu_z^{(i)}  \right)^{\top} \left( \bSigma_x^{(i)} \right)^{-1} \left( \bx^{(i)} - \bW  \bmu_z^{(i)}  \right) + \log \left| \bSigma_x^{(i)} \right|  \right\},
\end{equation}
ignoring any explicit reparameterization by $\beta$ for convenience.  If we optimize over $\bSigma_x^{(i)}$ in the feasible region $\bar{\calS}^{d}_{\alpha}$ and plug in the resulting value, then (\ref{eq:vae_cost_no_Sigma_z2_sup}) reduces to the new cost
\begin{equation} \label{eq:vae_cost_no_Sigma_z3_sup}
\calL(\bW,\bmu_z) \equiv \sum_{i,j}  \log \xi_{\alpha} \left( \left[ x_j^{(i)} - \bw_{j\cdot} \bmu_z^{(i)}   \right]^2 \right),
\end{equation}
an immaterial constant notwithstanding.  Given that $\lim_{t \rightarrow 0} \tfrac{1}{t}\left( |x|^t - 1 \right) = \log |x|$, and $\lim_{t \rightarrow 0} \sum_j |x_j|^p = \| \bx \|_0$, then up to an irrelevant scaling factor and additive constant, the stated result follows.

%=========================================================
\section{Proof of Theorem~\ref{thm:free_mu_x}}
\label{sec:proof_theorem3}

Based on the stated conditions, the VAE objective simplifies to
\begin{eqnarray} \label{eq:vae_cost_full_sup}
\calL(\btheta,\bphi)   =   \sum_i \left\{ \mathbb{E}_{q_{\tiny \bphi}\left(z|\bx^{(i)} \right)} \left[ \tfrac{1}{\lambda_x}\left\| \bx^{(i)} - \bmu_x(z;\btheta)  \right\|^2  \right]  + d \log \lambda_x + \lambda_z - \log \lambda_z    + \left( \ba^{\top} \bx^{(i)} \right)^2  \right\}.
\end{eqnarray}
Now choose some $\hat{\ba}$ such that $\hat{\mu}_z^{(i)} = \hat{\ba}^{\top} \bx^{(i)}$ has a unique value for every sample $\bx^{(i)}$ (here we assume that each sample is unique, although this assumption can be relaxed).  We then define the function $h : \mathbb{R} \rightarrow \{1,\ldots,n\}$ as
\begin{equation}
h(z) \triangleq \arg \min_{i \in \{1,\ldots,n\} } \| z - \mu_z^{(i)} \|_2.
\end{equation}
and the piecewise linear decoder mean function
\begin{equation}
\bmu_x(z; \btheta) = \bx^{(h(z))}.
\end{equation}
Given these definitions, (\ref{eq:vae_cost_full_sup}) becomes
\begin{eqnarray} \label{eq:vae_cost_full2_sup}
\calL(\btheta,\bphi)   =   \sum_i \left\{ \mathbb{E}_{q_{\tiny \bphi}\left(z|\bx^{(i)} \right)} \left[ \tfrac{1}{\lambda_x}\left\| \bx^{(i)} - \bx^{(h(z))}  \right\|^2  \right]  + d \log \lambda_x + \lambda_z - \log \lambda_z    + \left( \hat{\mu}^{(i)}_z \right)^2  \right\} \nonumber \\
=  \sum_i \left\{ \mathbb{E}_{p(\epsilon)} \left[ \tfrac{1}{\lambda_x}\left\| \bx^{(i)} - \bx^{(h[\hat{\mu}^{(i)}_z + \sqrt{\lambda_z} \epsilon])}  \right\|^2  \right]  + d \log \lambda_x + \lambda_z - \log \lambda_z    + \left( \hat{\mu}^{(i)}_z \right)^2  \right\}.
\end{eqnarray}
Now define the set
\begin{equation}
\calS^{(i)} \triangleq  \left\{\epsilon : h\left(\hat{\mu}^{(i)}_z + \sqrt{\lambda_z} \epsilon \right) = i \right\},
\end{equation}
which represents the set of $\epsilon$ that quantize to the correct index.  We then have
\begin{eqnarray} \label{eq:bound_full2_sup}
&& \hspace*{-1.0cm} \mathbb{E}_{p(\epsilon)} \left[ \tfrac{1}{\lambda_x}\left\| \bx^{(i)} - \bx^{(h[\hat{\mu}^{(i)}_z + \sqrt{\lambda_z} \epsilon])}  \right\|^2  \right]  \nonumber \\
& = & \int_{\epsilon \in \calS^{(i)} } \left[ \tfrac{1}{\lambda_x}\left\| \bx^{(i)} - \bx^{(h[\hat{\mu}^{(i)}_z + \sqrt{\lambda_z} \epsilon])}  \right\|^2  \right] p(\epsilon) d \epsilon + \int_{\epsilon \notin \calS^{(i)} } \left[ \tfrac{1}{\lambda_x}\left\| \bx^{(i)} - \bx^{(h[\hat{\mu}^{(i)}_z + \sqrt{\lambda_z} \epsilon])}  \right\|^2  \right] p(\epsilon) d \epsilon \nonumber \\
& = & \int_{\epsilon \notin \calS^{(i)} } \left[ \tfrac{1}{\lambda_x}\left\| \bx^{(i)} - \bx^{(h[\hat{\mu}^{(i)}_z + \sqrt{\lambda_z} \epsilon])}  \right\|^2  \right] p(\epsilon) d \epsilon \nonumber \\
& \leq & \int_{\epsilon \notin \calS^{(i)} }  \tfrac{\eta}{\lambda_x}  p(\epsilon) d \epsilon \\
& = & \tfrac{\eta}{\lambda_x} P\left( \epsilon \notin \calS^{(i)} \right), \nonumber
\end{eqnarray}
where
\begin{equation}
\eta \triangleq \max_{i,j \in \{1,\ldots,n\}, i \neq j } \| \bx^{(i)} - \bx^{(j)} \|_2^2,
\end{equation}
the maximal possible quantization error.  Now we also define
\begin{equation}
\rho \triangleq \max_{i,j \in \{1,\ldots,n\}, i \neq j } \frac{1}{2}\|  \hat{\mu}^{(i)}_z -  \hat{\mu}^{(j)}_z \|_2^2,
\end{equation}
which is half the minimum distance between any two $\hat{\mu}^{(i)}_z$ and $\hat{\mu}^{(j)}_z$, with $i \neq j$.  Then
\begin{eqnarray}
P\left( \epsilon \notin \calS^{(i)} \right) & \leq & P\left(\sqrt{\lambda_z}\epsilon > \rho \right) \nonumber \\
& \leq & \tfrac{\lambda_z}{\rho^2}
\end{eqnarray}
by Chebyshev's inequality as was used in proving Theorem \ref{thm:vae_and_rpca}.  This implies that (\ref{eq:vae_cost_full2_sup}) can be bounded via

\begin{eqnarray} \label{eq:vae_cost_full3_sup}
\calL(\btheta,\bphi)  & \leq &   \sum_i \left\{ \tfrac{\eta}{\lambda_x} P\left( \epsilon \notin \calS^{(i)} \right)  + d \log \lambda_x + \lambda_z - \log \lambda_z    + \left( \hat{\mu}^{(i)}_z \right)^2  \right\} \nonumber \\
& = &   \sum_i \left\{ \tfrac{\eta}{\rho^2}   + (d - 1) \log \alpha + \alpha + \left( \hat{\mu}^{(i)}_z \right)^2  \right\}
\end{eqnarray}
assuming we are at the trial solution $\hat{\lambda}_x = \hat{\lambda}_z = \alpha$.  As we allow $\alpha \rightarrow 0$, this expression is unbounded from below, and as an upper bound on the VAE objective, the theorem follows.  Incidentally, it should also be possible to prove that for $\alpha$ sufficiently small, no other solution can do appreciably better in terms of the dominate $(d-1) \log \alpha$ factor, but we will reserve this for future work.

\vskip 0.2in
\bibliography{refs}

\end{document}